\documentclass[10pt,twocolumn,letterpaper]{article}
\usepackage[pagenumbers]{cvpr} 

\definecolor{cvprblue}{rgb}{0.21,0.49,0.74}
\usepackage[pagebackref,breaklinks,colorlinks,allcolors=cvprblue]{hyperref}

\usepackage{booktabs}
\usepackage[table]{xcolor}   
\usepackage{multirow}
\definecolor{grpbg}{RGB}{235,240,250}
\newif\ifarxiv
\arxivtrue

\usepackage{tikz}
\usepackage{circledsteps}
\pgfkeys{/csteps/inner color=white}
\pgfkeys{/csteps/outer color=white}
\pgfkeys{/csteps/fill color=black}
\pgfkeys{/csteps/inner ysep=3pt}
\pgfkeys{/csteps/inner xsep=3pt}
\definecolor{mydarkblue}{rgb}{0,0.08,0.45}

\usepackage{booktabs}
\usepackage[most]{tcolorbox}

\newtcblisting{promptbox}[1]{
  enhanced,
  breakable,
  listing only,
  listing engine=listings,
  colback=gray!3,
  colframe=black!60,
  colbacktitle=black!70,
  coltitle=white,
  fonttitle=\bfseries\small,
  title={#1},
  arc=2pt,
  boxrule=0.5pt,
  left=7pt,
  right=7pt,
  top=5pt,
  bottom=5pt,
  before skip=8pt,
  after skip=10pt,
  listing options={
    basicstyle=\fontencoding{T1}\rmfamily\small,
    columns=fullflexible,
    keepspaces=true,
    breaklines=true,
    breakatwhitespace=false,
    breakindent=0pt,
    breakautoindent=false,
    showstringspaces=false,
    aboveskip=0pt,
    belowskip=0pt,
    literate={*}{{\char42}}1 {-}{{\char45}}1 {`}{{\char96}}1
  }
}

\def\confName{CVPR}
\def\confYear{2027}

\title{GlassGround: GUI Grounding Beyond Native Screenshots with \\ Egocentric Wearable Vision}

\author{
Yun Wang$^{1}$ \quad Siyue Sun$^{2}$ \quad Jiaying Du$^{1}$ \quad Xinyi Wei$^{1}$\\
Xiaoming Zhai$^{1}$ \quad Zhenlin An$^{1}$ \quad Ninghao Liu$^{3}$\\[2pt]
$^{1}$University of Georgia \quad $^{2}$Northeastern University \quad $^{3}$The Hong Kong Polytechnic University\\
{\tt\small \{yun.wang1, jd04294, xinyi.wei, xiaoming.zhai, zhenlin.an\}@uga.edu}\\
{\tt\small sun.siyue@northeastern.edu \quad ninghliu@polyu.edu.hk}
}

\begin{document}
\maketitle

\begin{abstract}
Wearable assistants must localize interface elements from camera observations when direct screenshot access is unavailable, yet existing GUI grounding benchmarks primarily evaluate native screenshots.
We introduce GlassGround, a new benchmark that pairs native screenshots with wearable camera observations while holding GUI states, instructions, and targets fixed.
GlassGround contains 3,250 tasks across Android and iOS, captured using four wearable devices under four viewing conditions.
Across ten open-source models, mean grounding accuracy drops from 94.8\% on screenshots to 48.5\% on camera images.
Further analysis reveals that changes in target scale, screen geometry, and background cannot fully explain this performance degradation.
We then study two ways to narrow this gap.
At inference time, local zooming and screen rectification help but leave most of the gap.
For training-time adaptation, we build on GlassGround's paired-view design and introduce PairDistill, a baseline that aligns target representations across views alongside coordinate supervision.
Trained on independently collected data and using no screenshots at inference, it improves camera-image accuracy by 9.4 and 12.2 points on two backbones and transfers to unseen devices and operating systems.
Combined with local zooming, it reaches 69.6\% and 64.4\%, but does not close the gap.
The GlassGround dataset and code are available at \url{https://github.com/yunnw/GlassGround}.
\end{abstract}

\section{Introduction}
\label{sec:intro}

\begin{figure}[t]
    \centering
    \includegraphics[width=\linewidth]{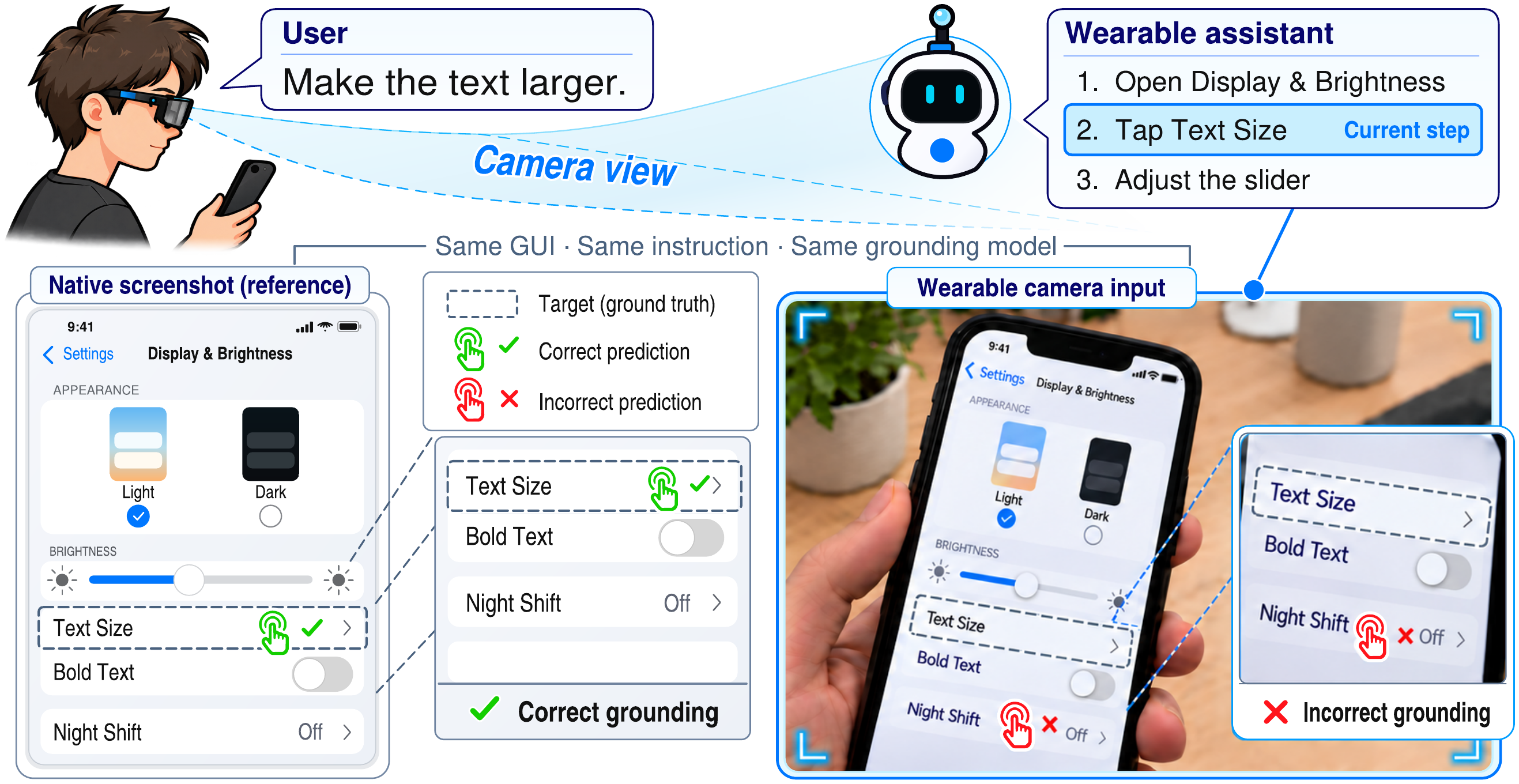}\vspace{-1mm}
    \caption{GUI grounding is more challenging in physical observations (e.g., wearable camera views) than in native screenshots.
    See our repository for a demo video of the live headset prototype.
    }
    \label{fig:teaser}\vspace{-3mm}
\end{figure}

Wearable AI assistants on smart glasses can guide users through smartphone tasks using egocentric camera views of the phone screen~\cite{li2025explorar}.
For example, when a user is learning an unfamiliar application, an assistant needs to identify the next action based on the current screen and update its guidance after the user acts~\cite{fang2026guideme}.
Making this guidance actionable requires visual GUI grounding, which localizes the interface element corresponding to an action instruction, typically in a native screenshot~\cite{cheng2024seeclick,li2025screenspot}.
However, unlike conventional GUI agents that operate directly on digital interfaces, wearable assistants may observe a smartphone only through their cameras, without direct access to its screenshots or underlying interface state~\cite{zhang2018robust}.
In these egocentric camera views, the interface may occupy only a small part of the frame and exhibit perspective distortion, while illumination, reflections, and display-camera imaging effects can reduce the visibility or legibility of its content~\cite{schmid2021screenshotmatcher,cheng2023recaptured}.

These visual changes may affect grounding: a model can correctly localize a target in a screenshot yet fail on the corresponding camera image (Fig.~\ref{fig:teaser}).
Evaluating physical GUI grounding therefore requires a benchmark that captures real camera observations while controlling for differences in task content.
However, existing GUI grounding benchmarks~\cite{cheng2024seeclick,li2025screenspot} primarily evaluate grounding on native screenshots, while camera-captured screen datasets~\cite{cheng2023recaptured,zhu2025comprehensive} focus on image restoration rather than instruction-conditioned grounding.

To address this need, we introduce \textbf{GlassGround}, a new benchmark for evaluating GUI grounding on real wearable camera observations.
It pairs native screenshots with camera observations of the same GUI state, instruction, and target.
To construct these pairs, we display each screenshot on its original smartphone and capture the interface using multiple wearable devices under varied viewing conditions.
Grounding tasks are constructed on native screenshots and shared across their paired camera observations to avoid the exclusion of targets that are difficult to discern in camera images.
This paired design enables us to compare the grounding performance of the same model on native screenshots and camera observations, and to analyze performance differences across devices and viewing conditions.
When evaluated on camera images, models receive only the image and instruction, without access to the paired screenshot.

Our evaluation using GlassGround reveals a \textbf{substantial physical grounding gap.}
Across ten open-source models, mean accuracy drops from 94.8\% on native screenshots to 48.5\% on physical observations, with considerable variation across devices and lighting conditions.
To investigate this gap, we construct intermediate observations that account for changes in target scale, screen geometry, and background.
Yet the remaining gap between these transformed screenshots and real camera images still accounts for nearly two-thirds of the original degradation.
We further evaluate \emph{ReGround}, a local crop-and-reground strategy to improve grounding on small targets~\cite{li2025screenspot,zhou2025mai}.
Although ReGround improves accuracy, it leaves the gap between transformed screenshots and physical observations largely unchanged.

Building on GlassGround's paired-view design, we further introduce \textbf{PairDistill}, which uses target representations from native screenshots to supervise corresponding representations in physical observations, alongside coordinate supervision. We construct training pairs by displaying annotated GUI screenshots on smartphones and capturing them with wearable cameras.
On GlassGround, PairDistill improves camera image grounding accuracy by 9.4 and 12.2~pp for UI-Venus-2-9B and MAI-UI-8B, respectively.
Combined with ReGround, their accuracies further increase to 69.6\% and 64.4\%.
These gains also generalize to unseen wearable devices, viewing conditions, and operating systems during training.
Our contributions are as follows.
\begin{itemize}
    \item We introduce GlassGround, a benchmark for controlled comparisons of GUI grounding on native screenshots and real wearable camera observations.

    \item We characterize the observation gap through intermediate observation stages, showing that most of the gap remains between transformed screenshots and real camera images.

    \item We show that inference-time processing and training-time adaptation (PairDistill) provide complementary gains, narrowing but not closing the physical grounding gap.
\end{itemize}

\begin{figure*}[t]
    \centering
    \includegraphics[width=\textwidth]{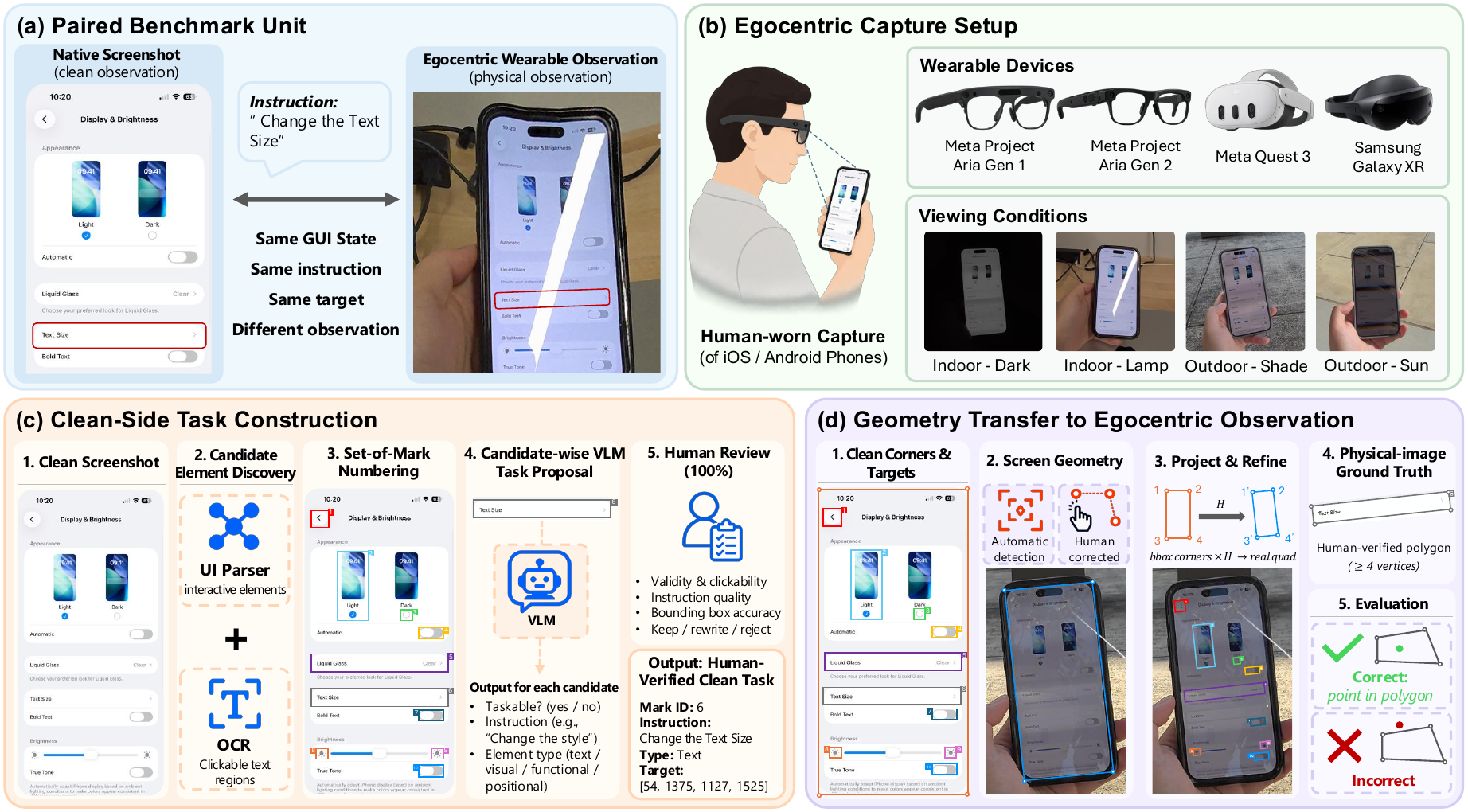}\vspace{-2mm}
    \caption{\textbf{Overview of the GlassGround benchmark construction.}
(a) Each benchmark unit pairs a native screenshot (\emph{Clean}) with an egocentric observation (\emph{Physical}) of the same GUI state, instruction, and target.
(b) Physical displays are captured with four head-worn devices under four viewing conditions.
(c) Tasks are constructed on clean screenshots using candidate discovery, VLM-based task proposals, and full human verification.
(d) Clean targets are transferred to egocentric images using human-verified screen geometry, homography, and optional local refinement, producing polygon ground truth for point-in-polygon evaluation.}\vspace{-3mm}
    \label{fig:overview}
\end{figure*}

\section{Related Work}

\subsection{Visual GUI Grounding and Benchmarks}
Visual GUI grounding localizes the interface element referred to by a natural language instruction and is a basic capability of GUI agents.
Grounding models such as UGround~\cite{gou2025navigating}, OS-Atlas~\cite{wu2025atlas}, and GUI-Actor~\cite{wu2026gui} learn this capability from large collections of GUI screenshots, and GUI agents such as Aguvis~\cite{xu2024aguvis} and UI-TARS~\cite{qin2025ui} build on it to operate interfaces from screenshots.
Grounding benchmarks have expanded alongside these models, starting with ScreenSpot~\cite{cheng2024seeclick} across mobile, desktop, and web interfaces.
ScreenSpot-Pro~\cite{li2025screenspot} extends evaluation to small targets in high-resolution professional software, OSWorld-G~\cite{xie2026scaling} and VenusBench-GD~\cite{zhou2025venusbench} cover more complex grounding tasks, and IGG~\cite{kimreveal} studies targets that become visible only after interaction.
Recently, ScreenHaystack~\cite{li-etal-2026-screenhaystack}, WinDeskGround~\cite{zhao2026windeskground}, and GUI-Perturbed~\cite{wang2026gui} analyze grounding robustness by controlling target position, window occlusion, interface layout and rendering, and instruction phrasing.
Whether they broaden task coverage or probe robustness, these benchmarks evaluate grounding on digital screenshots or rendered interfaces.
\textbf{GlassGround instead evaluates GUI grounding on physical screens observed through wearable cameras, introducing real display-to-camera imaging effects into the evaluation.}

\subsection{Camera-Observed and Physical Interfaces}
Prior work has examined camera-observed physical interfaces from several perspectives.
Screen recapture and demoiréing studies use the correspondence between source images and their photographs to analyze imaging degradation~\cite{yue2022recaptured,cheng2023recaptured,zhang2024character}, and PC-Restore~\cite{zhu2025comprehensive} further provides pairs of real camera-captured GUI images and clean references for image restoration with OCR-based evaluation.
ScreenshotMatcher~\cite{schmid2021screenshotmatcher} retrieves the native screenshot that corresponds to a photograph of a screen, and document rectification methods recover canonical geometry from photographs~\cite{ma2018docunet,das2019dewarpnet,feng2021doctr}.
Systems such as RoScript~\cite{qian2020roscript}, NiCro~\cite{xie2023nicro}, RoboTest~\cite{yu2024practical}, and BrushLens~\cite{liang2023brushlens} use external cameras to recognize, match, or operate elements on physical interfaces.
Egocentric studies address visual intention grounding~\cite{sun2025visual}, visual question answering on smart glasses~\cite{jiang2026superglasses}, pointing comprehension~\cite{li2026mllms}, and the transition from scene perception to web tasks~\cite{yu2026ego2web}.
Unlike prior camera-based interface work, \textbf{GlassGround benchmarks instruction-conditioned GUI grounding from wearable views, pairing camera observations with native screenshots to measure the grounding gap on identical tasks.}

\section{GlassGround: Benchmark Construction}
GlassGround is a benchmark with paired samples for GUI grounding from egocentric wearable viewpoints.
Each sample contains a native screenshot paired with egocentric images of the same interface shown on a physical smartphone and captured by wearable devices.
This design allows us to study the effect of physical observation on grounding with task content fixed.
Fig.~\ref{fig:overview} shows the construction process.

\subsection{Paired Task Formulation}
\label{sec:formulation}

A GlassGround task is defined as a triplet $(s, q, e)$: a GUI state $s$, an instruction $q$, and the target element $e \in \mathcal{E}(s)$ that $q$ refers to, where $\mathcal{E}(s)$ is the set of UI elements in $s$.
An observation $I$ is an image showing the interface in $s$.
For a given grounding task $(s, q, e)$, each observation yields an evaluation instance $(I, q, y)$, where $y$ is the region in $I$ corresponding to $e$.
Given an instance, a grounding model receives only $I$ and $q$ and predicts a point $\hat{p} \in \mathbb{R}^2$ in the coordinates of $I$.
The prediction is correct if $\hat{p} \in y$.

We consider two observation processes that produce different visual inputs for the same grounding task.
The \textbf{clean observation} $I_c = O_c(s)$ is the native screenshot of the GUI state.
The \textbf{physical observation} $I_p = O_p(s;\, \phi)$ is an egocentric image of the same state displayed on a physical screen and captured by a wearable device (see Fig.~\ref{fig:overview}(a)).
Here, $\phi$ denotes the capture setting.
For each GUI state, we obtain one clean observation and collect $K$ physical observations under different capture settings $\{\phi_k\}_{k=1}^{K}$, yielding $I_p^{(k)} = O_p(s;\, \phi_k)$.
Since they depict the same state, each $I_p^{(k)}$ is paired with $I_c$.
Each grounding task therefore yields $K+1$ instances, one per observation, and each physical instance is paired with the clean instance of the same task:
\begin{equation}
  (I_c, q, y_c) \longleftrightarrow
  \{(I_p^{(k)}, q, y_p^{(k)})\}_{k=1}^{K}.
\end{equation}
We refer to the change in the observation process, from $O_c$ to $O_p$ with the underlying state held fixed, as an \textbf{observation shift}.
It changes the model input from $I_c$ to $I_p^{(k)}$ while leaving the
task unchanged within each pair. 
This allows us to measure how physical observation affects grounding performance independently of differences in task content.

\subsection{Physical Display Capture Protocol}
\label{sec:capture}
We instantiate the physical observation process $O_p$ through a \textbf{capture protocol} designed to \textit{preserve identical GUI content while varying the conditions under which it is observed}. Specifically, we consider two major sources of variation: wearable imaging devices and viewing conditions. Their cross product defines $K=16$ capture settings, comprising four devices and four viewing conditions (Fig.~\ref{fig:overview}(b)).

\smallskip\noindent\textbf{Preserving GUI content.}\quad
Live interfaces may change across captures due to dynamic content.
To preserve an identical GUI state across all paired observations, we display each native screenshot $I_c$ full-screen on its original smartphone instead of repeatedly navigating to the same live interface.
All physical observations of a state thus share identical GUI content, so their differences from $I_c$ arise from the physical observation process.

\smallskip\noindent\textbf{Imaging devices.}\quad
Different wearable devices produce distinct image characteristics due to differences in optics, sensors, exposure control, and image signal processing (ISP)~\cite{Phan_2021_CVPR,Robidoux_2021_CVPR}.
To capture this variation, we include \emph{Meta Project Aria Gen~1} and \emph{Gen~2}, research egocentric glasses that record from onboard RGB cameras,
and \emph{Meta Quest~3} and \emph{Samsung Galaxy XR}, consumer XR headsets whose images are obtained through their built-in capture functions.

\smallskip\noindent\textbf{Viewing conditions.}\quad
We capture each GUI state under four lighting conditions that affect screen visibility and legibility~\cite{liu2014effect}:
\emph{Indoor-Dark} (room lights off),
\emph{Indoor-Lamp} (fixed indoor lighting),
\emph{Outdoor-Shade} (daylight without direct sunlight on the phone),
and \emph{Outdoor-Sun} (phone directly exposed to sunlight).

\smallskip\noindent\textbf{Handheld capture and quality control.}\quad
Operators wear the device and hold the phone naturally without a mount, allowing screen position, viewing distance, and viewing angle to vary within each capture setting.
Display settings remain fixed for each phone and viewing condition, with adaptive brightness disabled.
We retain device-specific appearance differences and natural artifacts such as mild blur, reflections, and partial occlusion, but retake captures if any target falls partially outside the image boundary (Supp.\ Sec.~\ref{supp:capture_protocol}).

\subsection{Grounding Task Construction}
\label{sec:paired_task}
We construct grounding tasks on native screenshots and reuse them across the corresponding physical observations.
For each state $s$, we define an instruction $q$ and target element $e$ from the clean screenshot $I_c$, together with the corresponding target region $y_c$ (Fig.~\ref{fig:overview}(c)).
We avoid using physical observations here, as doing so could preferentially omit tasks whose physical observations are hard to recognize.

To generate candidate tasks, we first identify potential target regions using OmniParser~\cite{lu2024omniparser}, supplemented by OCR to recover text regions that the parser may miss.
We then assign Set-of-Mark labels~\cite{yang2023set} to the candidates, and query a VLM (Claude Fable~5.1~\cite{fable51}) for each candidate individually.
It judges whether the candidate is a suitable target and, if so, writes a grounding instruction without predicting a target region.
It also classifies each instruction as \emph{text}, \emph{visual}, \emph{functional}, or \emph{positional} according to its primary grounding cue (see Supp.\ Sec.~\ref{supp:clean_annotation}).

Since generated instructions may contain ambiguous instructions or invalid targets, human annotators review every candidate.
They verify target validity and clickability, instruction clarity and type, and bounding-box accuracy.
Annotators may retain, revise, or reject proposals and add valid tasks missed by the automated pipeline. This yields the final $q$, $e$, and human-verified region $y_c$.
The resulting tasks are shared across all $K$ physical observations of each GUI state, with only their target regions transferred (Sec.~\ref{sec:gt_transfer}).

\subsection{Physical Ground-Truth Transfer}
\label{sec:gt_transfer}
To obtain target annotations for physical observations, we transfer the human-verified target regions from clean screenshots using geometric correspondence (Fig.~\ref{fig:overview}(d)).
Since the smartphone display is planar, a homography can map screenshot points into the wearable camera image.

For each physical observation $I_p^{(k)}$, we detect the screen corners using YOLO11n-pose~\cite{yolo11} fine-tuned on human-verified screen-corner annotations.
Since errors in screen localization would affect all transferred targets, annotators manually verify and correct every corner prediction.
Let $\{c_j\}_{j=1}^{4}$ denote the corners of $I_c$ and $\{c_j^{(k)}\}_{j=1}^{4}$ the corresponding verified screen corners in $I_p^{(k)}$.
We compute the homography $H_k$ determined by these correspondences,
\begin{equation}
  c_j^{(k)}
  =
  \pi\!\left(H_k\tilde{c}_j\right),
  \qquad j=1,\ldots,4,
\end{equation}
where $\tilde{c}_j$ denotes the homogeneous coordinates of $c_j$, and $\pi(\cdot)$ maps them back to image coordinates.
Applying the homography to the clean target region, with mesh refinement where local misalignment remains, yields the ground-truth region $y_p^{(k)}$, which we retain as a polygon to preserve its perspective geometry.
Cross-review of all images shifts accuracy by at most 0.16~pp (see Supp.\ Sec.~\ref{supp:physical_annotation}).

\subsection{Benchmark Statistics}
\label{sec:bench_stats}
GlassGround contains 536 GUI states from 102 applications across iOS and Android, comprising 303 iOS states and 233 Android states, with 3,250 grounding tasks in total.
The applications span eight categories:
\emph{System}, \emph{Media}, \emph{Work}, \emph{Shopping},
\emph{Travel}, \emph{Lifestyle}, \emph{Social}, and \emph{Web}.
Category, application, and instruction-type distributions are provided in Supp.\ Sec.~\ref{supp:clean_annotation}.
Each GUI state is captured under the full $4\times4$ cross product of wearable devices and viewing conditions, yielding 8,576 unique physical images.
Because multiple grounding tasks may share the same GUI state, these captures correspond to 52,000 task--observation instances.
Thus, each of the 3,250 tasks forms one clean observation paired with 16 physical observations.

\definecolor{baselinebg}{RGB}{250,240,239}
\definecolor{oursbg}{RGB}{229,239,250}

\begin{table*}[t]
\centering
\caption{
Grounding accuracy (\%) on GlassGround.
Single-pass results cover observation stages and rectification with annotated (Oracle) or detected (Auto) screen corners.
ReGround is reported separately.
Open-source baselines are grouped into GUI-specialized and general-purpose models.
Each column highlights the \textbf{best} and \underline{second-best} scores on the full benchmark.
$^\dagger$ Evaluated on a shared stratified subset.
}\vspace{-2mm}
\label{tab:main_results}
\footnotesize
\setlength{\tabcolsep}{0pt}
\begin{tabular*}{\textwidth}{@{\extracolsep{\fill}} l cccc cc ccc}
\toprule
\multirow{2}{*}{\textbf{Model}}
 & \multicolumn{4}{c}{\textbf{Single-pass Accuracy} $\uparrow$}
 & \multicolumn{2}{c}{\textbf{Rectified (Single-pass)} $\uparrow$}
 & \multicolumn{3}{c}{\textbf{ReGround Accuracy} $\uparrow$} \\
\cmidrule(lr){2-5}\cmidrule(lr){6-7}\cmidrule(lr){8-10}
 & Clean & Resized & Warped & Physical & Oracle & Auto
 & Resized & Warped & Physical \\
\midrule

\rowcolor{baselinebg}
\multicolumn{10}{c}{\textit{Closed-source VLMs}$^\dagger$} \\

Claude Sonnet 5.5~\cite{anthropic2026sonnet55}
 & 97.20 & 90.13 & 94.95 & 70.29 & 71.49 & 69.96
 & 96.51 & 96.98 & 76.83 \\

GPT-6 Luna~\cite{openai2026gpt6}
 & 95.80 & 86.56 & 88.96 & 58.35 & 47.53 & 46.66
 & 90.39 & 91.44 & 61.28 \\

Jev 1.14~\cite{typesafe2026jev114}
 & 83.60 & 31.03 & 27.14 & 16.53 & 48.91 & 49.18
 & 49.76 & 48.28 & 31.35 \\

\midrule
\rowcolor{baselinebg}
\multicolumn{10}{c}{\textit{Open-source baselines}} \\

MAI-UI-8B~\cite{zhou2025mai}
 & 95.45 & 84.12 & 75.41 & 45.80 & 62.54 & 62.63
 & 92.43 & 89.75 & 59.03 \\

GUI-Owl-1.5-8B~\cite{xu2026mobile}
 & 92.37 & 81.72 & 74.29 & 44.28 & 56.33 & 56.43
 & 88.89 & 85.73 & 53.62 \\

GUI-Owl-1.5-32B~\cite{xu2026mobile}
 & 95.17 & 83.74 & 77.06 & 49.07 & 60.34 & 60.30
 & 90.81 & 88.21 & 57.09 \\

KV-Ground-8B~\cite{kvground2026}
 & 94.15 & 83.63 & 77.41 & 47.93 & 60.17 & 60.09
 & 90.23 & 87.06 & 56.16 \\

UI-Venus-2-9B~\cite{team2026ui}
 & 95.94 & 87.92 & \underline{84.43} & 55.13 & 65.97 & 65.82
 & 93.66 & 91.77 & 63.68 \\

UI-Mate-27B~\cite{ding2026ui}
 & \textbf{96.46} & 85.69 & 80.97 & 49.77 & 64.64 & 64.53
 & 93.03 & 91.32 & 59.93 \\

Holo3.1-35B-A3B~\cite{hai2026holo31}
 & 94.52 & 88.36 & 78.20 & 46.78 & 63.63 & 63.59
 & 93.32 & 90.54 & 59.69 \\

Indeed-UI-32B~\cite{indeedui2026}
 & 95.78 & 85.12 & 78.71 & 50.28 & 60.68 & 60.47
 & 91.87 & 89.32 & 57.57 \\

\midrule[0.3pt]

Muse Glimmer-30B~\cite{museglimmer2026}
 & 95.97 & 85.16 & 67.26 & 39.76 & \underline{69.50} & \underline{69.18}
 & 93.78 & \underline{92.43} & 62.88 \\

Qwen3.8-27B~\cite{qwen38}
 & 92.15 & \underline{88.65} & 83.52 & 56.04 & 67.52 & 67.37
 & \textbf{94.16} & 92.21 & 64.13 \\

\midrule
\textbf{Open-source Avg.}
 & 94.80 & 85.41 & 77.73 & 48.48 & 63.13 & 63.04
 & 92.22 & 89.83 & 59.38 \\

\midrule
\rowcolor{oursbg}
\multicolumn{10}{c}{\textit{Adapted models (mean $\pm$ std., 3 seeds)}} \\

UI-Venus-2-9B + \textbf{PD}
 & $\underline{96.24}_{\pm 0.37}$
 & $\mathbf{89.90}_{\pm 0.23}$
 & $\mathbf{87.80}_{\pm 0.27}$
 & $\mathbf{64.48}_{\pm 0.53}$
 & $\mathbf{69.58}_{\pm 0.58}$
 & $\mathbf{69.65}_{\pm 0.62}$
 & $\underline{93.91}_{\pm 0.17}$
 & $\mathbf{92.86}_{\pm 0.18}$
 & $\mathbf{69.62}_{\pm 0.47}$ \\

MAI-UI-8B + \textbf{PD}
 & $94.74_{\pm 0.16}$
 & $85.46_{\pm 0.70}$
 & $82.33_{\pm 0.38}$
 & $\underline{57.96}_{\pm 0.21}$
 & $65.20_{\pm 0.45}$
 & $65.17_{\pm 0.41}$
 & $91.53_{\pm 0.33}$
 & $90.09_{\pm 0.46}$
 & $\underline{64.36}_{\pm 0.10}$ \\

\bottomrule
\end{tabular*}\vspace{-3mm}
\end{table*}

\section{Analyzing the Physical Grounding Gap}
\label{sec:physical_gap}

We use GlassGround to evaluate GUI grounding in physical environments.
Our analysis addresses three questions:
(1) How large is the gap between native screenshots and physical observations, and how does it vary across models and capture settings?
(2) To what extent can changes in target scale, screen geometry, and background explain this degradation?
(3) To what extent can inference-time processing recover grounding accuracy without changing the model?

\subsection{Evaluation Setup}
\label{sec:evaluation_setup}

\noindent\textbf{Models.}\quad
Our model selection prioritizes recent GUI-specialized models with strong reported performance on screenshot-based GUI grounding benchmarks, allowing us to examine whether their grounding capabilities transfer to physical observations.
Our open-source suite includes eight \textit{GUI-specialized models}:
MAI-UI-8B~\cite{zhou2025mai}, GUI-Owl-1.5-8B and GUI-Owl-1.5-32B~\cite{xu2026mobile}, KV-Ground-8B~\cite{kvground2026},
UI-Venus-2-9B~\cite{team2026ui}, UI-Mate-27B~\cite{ding2026ui}, Holo3.1-35B-A3B~\cite{hai2026holo31}, and Indeed-UI-32B~\cite{indeedui2026}.
To broaden the comparison beyond GUI-specialized models, we evaluate two \textit{general-purpose VLMs}, Muse Glimmer-30B~\cite{museglimmer2026} and Qwen3.8-27B~\cite{qwen38}.
For \textit{closed-source baselines}, we consider inference efficiency for wearable assistance and select
Claude Sonnet 5.5~\cite{anthropic2026sonnet55} and GPT-6 Luna~\cite{openai2026gpt6}, two efficiency-oriented VLMs,
and Jev 1.14~\cite{typesafe2026jev114}, which targets fast structured prediction.
Open-source models are evaluated on the full benchmark, while closed-source models use a fixed subset comprising 500 clean and 8,000 physical instances.

\smallskip\noindent\textbf{Inference and evaluation.}\quad
By default, the model predicts the target location from the input image and instruction in a single call (\emph{Single-pass}).
In physical observations, the screen occupies only part of the image, so target elements occupy a smaller fraction of it than in clean observations.
We therefore also evaluate \emph{ReGround}, following prior work showing that region-focused refinement can improve small-target GUI grounding~\cite{li2025screenspot,zhou2025mai,luo2025visual,park2025r}.
ReGround re-localizes the target in an enlarged crop around the initial prediction, using at most one additional model call.
For evaluation, we use the predicted point or the center of the predicted bounding box, depending on the model's output format.
Grounding accuracy is the percentage of evaluated instances satisfying the point-in-polygon criterion.
Further details and matched-subset comparisons are in Supp.\ Sec.~\ref{supp:evaluation_details}.

\begin{figure*}[t]
    \centering
    \includegraphics[width=\textwidth]{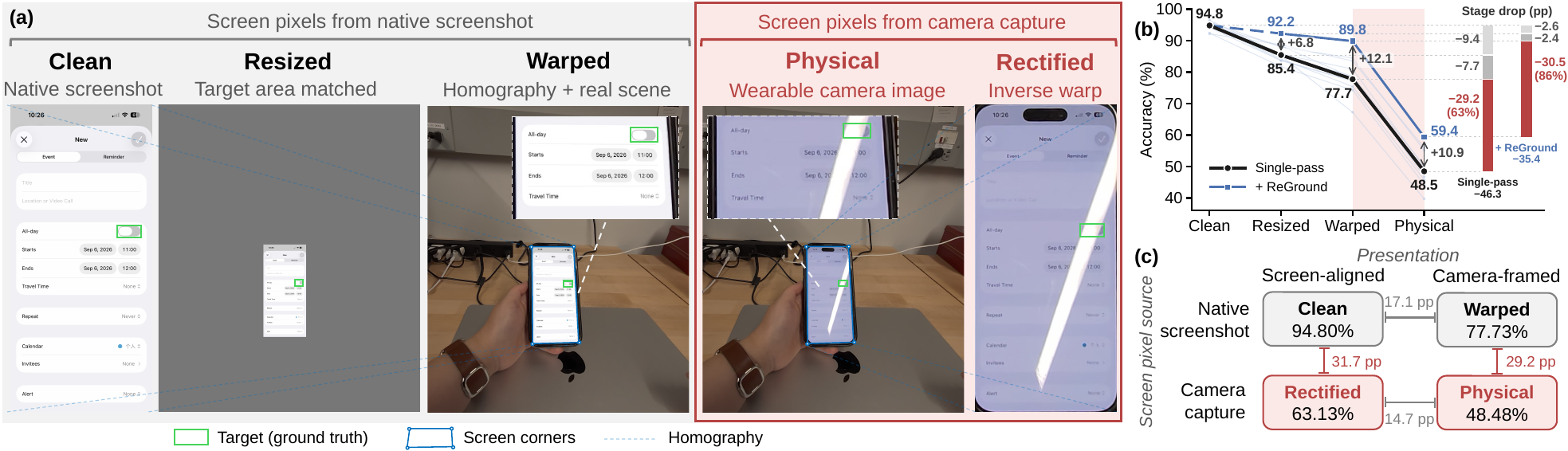}\vspace{-3mm}
    \caption{\textbf{Controlled analysis.} (a)~Different observation stages.
    \emph{Clean} is the native screenshot.
    \emph{Resized} scales the screenshot to match the target area in the corresponding physical image.
    \emph{Warped} maps the screenshot to the annotated screen while retaining the surrounding scene.
    \emph{Physical} is the original wearable camera image.
    (b)~Grounding accuracy across observation stages, with and without ReGround.
    (c)~Accuracy differences associated with screen pixel source and presentation.}
    \vspace{-4mm}
    \label{fig:observation_stages}
\end{figure*}

\begin{figure}[t]
    \centering
    \includegraphics[width=\linewidth]{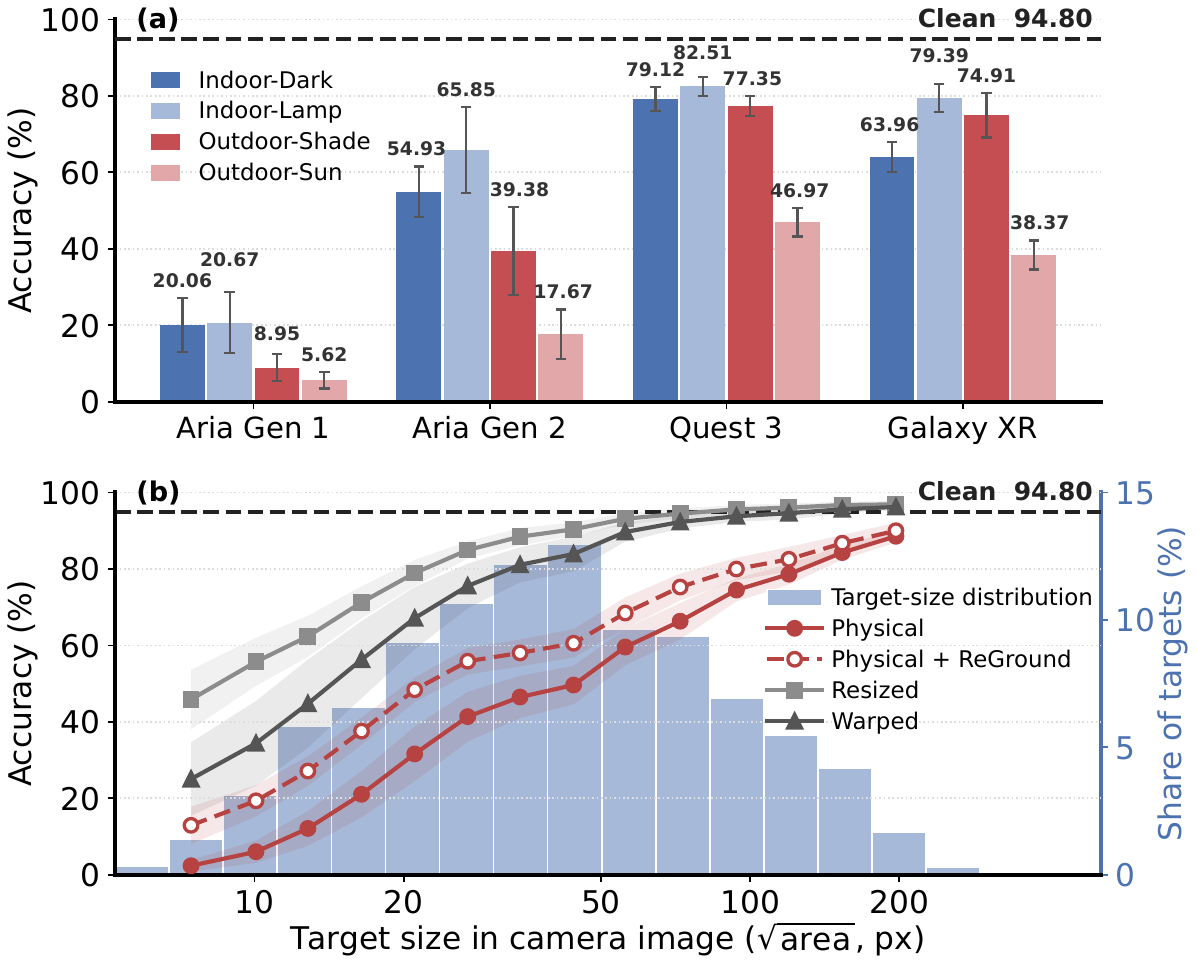}\vspace{-4mm}
    \caption{Variation of the physical observation gap.
    (a) Single-pass accuracy on Physical across devices and viewing conditions.
    (b) Accuracy by target size.
    Results are averaged over ten open-source models, with shading indicating standard deviations.
    }\vspace{-4mm}
    \label{fig:device_lighting}
\end{figure}

\subsection{Evaluating GUI Grounding with GlassGround}
\label{sec:gap_magnitude}
We evaluate GUI grounding models on GlassGround. Our analysis focuses on three aspects: overall grounding accuracy, performance differences across models, and variations across wearable devices and viewing conditions. 
Results are reported in Table~\ref{tab:main_results}. Some observations are as below.

\smallskip\noindent \Circled{\footnotesize 1} \textbf{Existing GUI grounding models struggle with physical observations.}\quad
All ten open-source models achieve an average accuracy of 94.8\% on native screenshots, but only 48.5\% on physical observations of the same tasks. The gap is consistent across models, with individual drops ranging from 36.1 to 56.2 percentage points (pp). Three closed-source models show similar degradation (26.9–67.1 pp). These results highlight the limited transfer of existing screenshot grounding capabilities to physical environments.

\smallskip\noindent \Circled{\footnotesize 2} \textbf{High screenshot accuracy does not guarantee strong physical performance.}\quad
Muse Glimmer-30B and UI-Venus-2-9B achieve similar accuracy on Clean ($\sim$96\%), but differ substantially on Physical (39.8\% vs. 55.1\%). Across the ten open-source models, the accuracy range widens from 4.3 pp on Clean to 16.3 pp on Physical. Model rankings also change considerably: Qwen3.8-27B moves from last on Clean to first on Physical, while Muse Glimmer-30B falls from second to last. These results show that strong performance on screenshot-based benchmarks does not necessarily translate into strong physical performance.

\smallskip\noindent \Circled{\footnotesize 3} \textbf{The physical grounding gap varies widely across devices and viewing conditions.}\quad
Averaged over models and conditions (Fig.~\ref{fig:device_lighting}(a)), grounding accuracy is highest on Quest 3, followed by Galaxy XR, Aria Gen 2, and Aria Gen 1.
Within each device, Outdoor-Sun is the hardest condition and Indoor-Lamp the easiest.
The relative difficulty of viewing conditions varies by device: Galaxy XR performs better under Outdoor-Shade than Indoor-Dark (74.9\% vs.\ 64.0\%), whereas the other devices show the opposite trend.
Imaging effects may contribute to these patterns: some Galaxy XR Indoor-Dark images lose fine text detail, while reflections and screen banding may help explain why Outdoor-Sun is the hardest condition across devices (Supp.\ Fig.~\ref{fig:supp_degradation_examples}).
Claude Sonnet 5.5 shows the same patterns, but reaches 59--66\% on Aria Gen~1 indoors, where open-source models average 20\% (Supp.\ Tab.~\ref{tab:supp_closed_conditions}).
These results highlight the importance of \textit{evaluating physical GUI grounding across diverse devices and viewing conditions}.

\subsection{Physical Grounding Gap: Controlled Analysis}
\label{sec:gap_decomposition}

To better understand the physical grounding gap, we investigate how different visual changes affect grounding performance. Specifically, we construct intermediate observations between native screenshots and physical camera images (Fig.~\ref{fig:observation_stages}(a)): (1) \textit{Resized} scales the native screenshot to match the target area in Physical; (2) \textit{Warped} introduces perspective distortion and real backgrounds by projecting the screenshot using a homography estimated from human-verified corners. In contrast, \textit{Rectified} applies an inverse homography, producing a front-facing view while retaining camera-captured screen content.


\smallskip\noindent \Circled{\footnotesize 4} \textbf{The observation gap remains between transformed screenshots and real camera images.}\quad
Across the ten open-source models, mean accuracy decreases from 94.8\% on Clean to 85.4\% on Resized, 77.7\% on Warped, and 48.5\% on Physical (Fig.~\ref{fig:observation_stages}(b)). The Warped-to-Physical gap is 29.2 pp, which is much larger than drops of 9.4 pp (Clean to Resized) and 7.7 pp (Resized to Warped).
The same pattern holds across models, with Warped-to-Physical drops ranging from 27.5 to 31.4 pp.
The gap is also larger for smaller targets, reaching approximately 35 pp at 20 px, compared with 8 pp at 200 px (Fig.~\ref{fig:device_lighting}(b)).
These findings suggest that changes in target scale, screen geometry, and background \textit{do not fully explain the physical grounding gap}.

\smallskip\noindent \Circled{\footnotesize 5} \textbf{Camera-captured screen is a larger contributor to the physical grounding gap than screen presentation.}\quad
We further compare Clean, Warped, Rectified, and Physical along two dimensions.
As shown in Fig.~\ref{fig:observation_stages}(c), changing the screen pixel source from native screenshots to camera captures corresponds to accuracy gaps of 29.2-31.7 pp. In contrast, changing screen presentation while retaining the same pixel source produces \textit{smaller gaps} of 14.7-17.1 pp.

\subsection{Can Inference-Time Methods Bridge the Gap?}
\label{sec:deployment_variation}

Finally, we examine whether inference-time methods can improve physical grounding, considering ReGround (Sec.~\ref{sec:evaluation_setup}) and screen rectification.

\smallskip\noindent \Circled{\footnotesize 6} \textbf{ReGround improves accuracy but still leaves a gap.}\quad
According to Fig.~\ref{fig:observation_stages}(b), ReGround raises mean accuracy on Resized and Warped to 92.2\% and 89.8\%, respectively, narrowing the Clean-to-Warped gap from 17.1 to 5.0 percentage points (pp). 
It also improves Physical accuracy from 48.5\% to 59.4\%. 
However, the Warped-to-Physical gap remains \textit{nearly unchanged} (29.2 vs. 30.5 pp), suggesting that local zooming does not address the main difficulties associated with camera-captured screen content.

\smallskip\noindent \Circled{\footnotesize 7} \textbf{Single-pass grounding after rectification can outperform ReGround.}\quad
Since human-verified screen corners (Oracle) are unavailable at inference time, we rectify with corners predicted by the detector in Sec.~\ref{sec:gt_transfer} (Auto).
With a single grounding call, Auto outperforms ReGround (63.0\% vs.\ 59.4\% on average) and approaches Oracle accuracy (63.1\%).
Auto remains close to Oracle on held-out states (Supp.\ Sec.~\ref{supp:auto_rectification}).
Yet rectification retains camera-captured screen content, and Oracle remains 31.7~pp below Clean (Fig.~\ref{fig:observation_stages}(c)).
Neither inference-time method bridges the gap to Clean, motivating model adaptation in the next section.


\section{Learning from Paired GUI Observations}
\label{sec:gap_mitigation}

The previous analysis reveals that existing GUI grounding models struggle with camera-captured screen content, even after inference-time processing.
Building on GlassGround's \emph{paired-view design}, we present \textbf{PairDistill}, which explores leveraging corresponding native screenshots and physical observations to improve grounding through representation alignment.
Specifically, PairDistill aligns representations of the same GUI target across the two views while supervising target prediction. Native screenshots provide supervision only during training. At inference, the adapted model takes only the physical image and task instruction.


\subsection{Reusing Screenshot Grounding Supervision}
\label{sec:paired_training_setup}

PairDistill is trained on paired examples, each consisting of a native screenshot $I_c$ and a physical image $I_p$ of the same GUI state, a task instruction $q$, and an answer $a$.
Here, $a$ denotes the target coordinates in $I_p$, formatted according to the backbone's output convention.
Representation alignment additionally requires the corresponding \textbf{target regions} $R_c$ and $R_p$ in the two views.
Following GlassGround's paired-view construction, we obtain such training examples by displaying GUI screenshots on smartphones and capturing them with wearable cameras. The original target annotation provides $R_c$. We then use the screen homography (Sec.~\ref{sec:gt_transfer}) to transfer $R_c$ into the physical image, obtaining $R_p$ and the corresponding answer $a$.

\subsection{PairDistill: Target-Aligned Distillation}
\label{sec:paired_adaptation}

\begin{figure}[t]
    \centering
    \includegraphics[width=\linewidth]{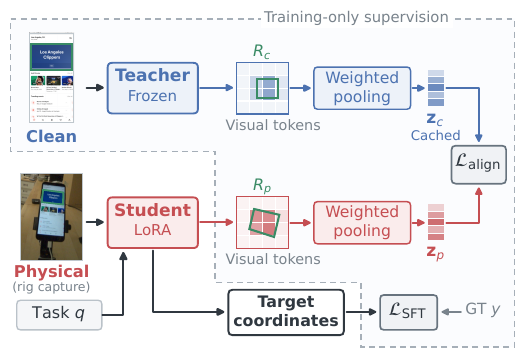}\vspace{-4mm}
    \caption{The overall framework of PairDistill.
    }\vspace{-5mm}
    \label{fig:pairdistill}
\end{figure}

PairDistill combines coordinate supervision with target representation alignment, as illustrated in Fig.~\ref{fig:pairdistill}. The former trains the model to predict target coordinates from physical observations $I_p$, while the latter uses clean screenshots $I_c$ to guide the representations of the corresponding targets.


\smallskip\noindent\textbf{Teacher and student.}\quad
The teacher and student are initialized from the same pretrained checkpoint.
The frozen teacher processes native screenshots, while the student takes physical images and task instructions as input.
We adapt the student using LoRA and extract visual token features $F_c$ and $F_p$ from the patch merger outputs of the teacher and student, respectively.
Since the teacher's features do not depend on the task instruction, we precompute and cache them.

\smallskip\noindent\textbf{Target pooling.}\quad
The same GUI target may occupy different positions and cover different numbers of visual tokens across the two views.
We therefore pool visual tokens based on their overlap with the target region.
Let $F=\{\mathbf{f}_j\}$ denote the visual token features of an image and $\Omega_j$ the image region covered by token $j$.
For a target region $R$, we define the pooled target feature $\operatorname{Pool}_R(F)$ as the area-weighted average of the token features:
\vspace{-3pt}
\begin{equation}
  \operatorname{Pool}_R(F)=\sum_j w_j\,\mathbf{f}_j,
  \quad
  w_j=\frac{|R\cap\Omega_j|}{\sum_k |R\cap\Omega_k|},
  \label{eq:target_pooling}
\vspace{-3pt}
\end{equation}
where $|\cdot|$ denotes area.
These area weights reduce the contribution of tokens that overlap only a small portion of the target.
We subtract the means $\bar{\mathbf{f}}_c$ and $\bar{\mathbf{f}}_p$ of all visual token features in each image from the pooled target features, obtaining the corresponding target representations
$\mathbf{z}_c=\operatorname{Pool}_{R_c}(F_c)-\bar{\mathbf{f}}_c$ and
$\mathbf{z}_p=\operatorname{Pool}_{R_p}(F_p)-\bar{\mathbf{f}}_p$.

\smallskip\noindent\textbf{Training objective.}\quad
We combine the two forms of supervision in the following objective:
\begin{equation}
  \mathcal{L}_{\mathrm{PD}}
  =\mathbb{E}_{\mathcal{D}}\!\left[\mathcal{L}_{\mathrm{SFT}}(I_p,q,a)+\lambda\,\mathcal{L}_{\mathrm{align}}\right],
  \label{eq:pairdistill}
\end{equation}
where $\mathcal{D}$ is the paired training set, $\mathcal{L}_{\mathrm{SFT}}$ is the mean token cross-entropy over $a$ with all other tokens masked, and $\mathcal{L}_{\mathrm{align}}=1-\cos(\mathbf{z}_p,\mathbf{z}_c)$.
We set $\lambda=1$.
Setting $\lambda=0$ yields the coordinate-only SFT baseline.


\section{Adaptation Results}
\label{sec:adaptation_results}


\smallskip\noindent\textbf{Experimental setup.}\quad
Following Sec.~\ref{sec:paired_training_setup}, we construct training data independently of GlassGround from Android screenshots in AMEX~\cite{chai2025amex}, displayed on a smartphone and recaptured with a fixed rig.
The training set contains 4,636 image pairs,
covering 76,224 targets across four device and lighting combinations: Aria Gen~1 under Indoor-Lamp and Indoor-Dark, and Quest~3 under Indoor-Dark and Outdoor-Sun.
We adapt UI-Venus-2-9B~\cite{team2026ui} and MAI-UI-8B~\cite{zhou2025mai} using LoRA on both the language model and vision encoder, including the patch merger, so that $\mathcal{L}_{\mathrm{align}}$ can update visual representations.
Following WiSE-FT~\cite{wortsman2022robust}, we apply weight interpolation with a fixed $\alpha=0.5$.
Checkpoints are selected on a screenshot-disjoint AMEX validation split, and all adapted results are averaged over three seeds.
By default, PairDistill uses scale augmentation during training (ablated in Table~\ref{tab:adaptation_ablation}).
Further details are in Supp.\ Sec.~\ref{supp:pairdistill_details}.

\subsection{Effectiveness of Paired Supervision}
\label{sec:adaptation_effectiveness}

\smallskip\noindent \Circled{\footnotesize 1} \textbf{PairDistill improves accuracy on physical images without additional inference-time processing.}\quad
Table~\ref{tab:main_results} compares PairDistill with baseline models (Base).
Under Single-pass inference, PairDistill improves Physical accuracy from 55.1\% to 64.5\% for UI-Venus-2-9B and from 45.8\% to 58.0\% for MAI-UI-8B.
Meanwhile, Clean accuracy stays within 0.8~pp of Base, indicating that \textbf{adaptation largely preserves grounding on native screenshots}.

\smallskip\noindent \Circled{\footnotesize 2} \textbf{PairDistill narrows the gap between transformed screenshots and physical observations.}\quad
The Warped-to-Physical gap decreases from 29.3 to 23.3 pp for UI-Venus and from 29.6 to 24.4 pp for MAI-UI. This contrasts with ReGround, which leaves this gap nearly unchanged (Sec.~\ref{sec:deployment_variation}).
Nevertheless, the remaining gap indicates that physical observations still pose substantial challenges.

\smallskip\noindent \Circled{\footnotesize 3} \textbf{PairDistill is complementary to inference-time methods.}\quad
Combining ReGround with PairDistill raises Physical accuracy to 69.6\% for UI-Venus and 64.4\% for MAI-UI, further improving over PairDistill alone by 5.1 and 6.4 pp.
Stage-wise analysis shows that ReGround mainly reduces the Clean-to-Resized gap, whereas PairDistill reduces the gaps between Resized and Physical (see Supp.\ Sec.~\ref{supp:stagewise}).
Auto rectification gives similar results (69.7\% and 65.2\%), 3.8 and 2.5~pp above Base under the same rectification.

\smallskip\noindent \Circled{\footnotesize 4} \textbf{Both coordinate supervision and target alignment contribute.}\quad
Without scale augmentation (Table~\ref{tab:adaptation_ablation}), coordinate supervision alone (SFT) raises accuracy from 55.1\% to 60.5\% for UI-Venus and from 45.8\% to 53.4\% for MAI-UI, showing that pairing enables effective reuse of screenshot annotations for physical grounding.
Under matched settings, adding target alignment yields further gains of 3.6 and 3.7~pp without scale augmentation and 2.7 and 3.7~pp with it.
Scale augmentation provides smaller gains (0.4--1.3~pp), and the full configuration achieves the highest accuracy on both backbones (further analyses in Supp.\ Sec.~\ref{supp:additional_adaptation}).

\begin{table}[t]
\centering
\footnotesize
\setlength{\tabcolsep}{2.4pt}
\caption{
Ablations on Physical under Single-pass, mean$_{\pm\text{std}}$ over three seeds.
PD: PairDistill; Aug: scale augmentation.
}\vspace{-4mm}
\label{tab:adaptation_ablation}
\begin{tabular}{lccccc}
\toprule
 & Base & SFT & SFT+Aug & PD & PD+Aug \\
\midrule
UI-Venus & 55.13 & $60.49_{\pm1.16}$ & $61.83_{\pm0.39}$ & $64.07_{\pm0.16}$ & $\mathbf{64.48}_{\pm0.53}$ \\
MAI-UI   & 45.80 & $53.44_{\pm2.01}$ & $54.30_{\pm1.84}$ & $57.17_{\pm0.59}$ & $\mathbf{57.96}_{\pm0.21}$ \\
\bottomrule
\end{tabular}\vspace{-4mm}
\end{table}
\subsection{Generalization Across Deployment Settings}
\label{sec:adaptation_generalization}

We examine whether PairDistill generalizes beyond its training conditions. While its training uses Android interfaces captured with a fixed rig, GlassGround evaluates natural handheld observations across diverse devices, lighting conditions, and operating systems.
The results are Single-pass accuracies on Physical, compared with Base.

\smallskip\noindent \Circled{\footnotesize 5} \textbf{PairDistill generalizes to unseen devices and lighting conditions.}\quad
Both backbones outperform Base across all 16 device-lighting combinations (Supp.\ Fig.~\ref{fig:adaptation_generalization}).
On the two unseen devices, Aria Gen 2 and Galaxy XR, PairDistill improves accuracy from 62.4\% to 71.0\% for UI-Venus and from 49.5\% to 63.4\% for MAI-UI.
These gains of 8.7 and 13.9~pp are comparable to or larger than the overall gains of 9.4 and 12.2~pp.
Across the 12 device-lighting combinations absent from training, the gains are 8.3 and 11.8~pp
over Base and 2.3 and 3.3~pp over SFT with matched augmentation.
Under Outdoor-Shade, which never appears in training, both backbones also improve on all four devices.

\smallskip\noindent \Circled{\footnotesize 6} \textbf{The improvements extend to unseen operating systems and applications.}\quad
Although training uses only Android interfaces, PairDistill improves iOS accuracy by 8.9 and 12.1 pp for UI-Venus and MAI-UI, respectively, comparable to their Android gains of 10.0 and 12.3 pp.
On iOS, it also outperforms matched SFT by 2.5 and 3.5~pp.
After removing all applications whose names also appear in AMEX, including browsers, app stores, and system applications, the remaining subset contains 7,600 physical images and 46,432 instances.
In this subset, the gains are 9.2 and 12.0~pp, closely matching those of the full benchmark.

\section{Conclusion and Limitations}
We presented GlassGround, a paired benchmark that holds GUI states, instructions, and targets fixed to study grounding across native screenshots and wearable camera observations.
Our analysis reveals substantial losses when moving from transformed screenshot content to real camera images, even with local zooming.
We further introduced PairDistill, which uses target correspondences between paired views to supervise grounding on physical images.
Its gains extend to unseen wearable devices, capture configurations, and operating systems without paired screenshots at inference.
These results highlight observation conditions as an important dimension of GUI grounding evaluation and demonstrate the value of paired supervision for adaptation.

Our benchmark focuses on single-step grounding in static images and does not evaluate complete interaction sequences.
The evaluation is also limited to smartphone screens, and performance on other physical displays remains to be assessed.
Attributing the Warped-to-Physical gap to specific imaging factors is also left to future work.

\section*{LLM Usage Statement}
Claude Fable 5.1 was used during benchmark construction to assess candidate GUI elements, generate grounding instructions, and classify instruction types.
It did not produce target regions.
All retained tasks, instructions, and target annotations were reviewed and verified by human annotators.
OpenAI ChatGPT and Anthropic Claude were used to assist with manuscript organization, wording, and editorial feedback.
The authors reviewed all AI-assisted content and take full responsibility for the final manuscript.

{
    \small

}


\clearpage
\ifarxiv\else\setcounter{page}{1}\fi
\maketitlesupplementary

\section{GlassGround Benchmark Details}
\label{supp:benchmark_details}

\subsection{Physical Capture Setup and Protocol}
\label{supp:capture_protocol}

\smallskip\noindent\textbf{Source smartphones and display settings.}\quad
GlassGround uses three smartphones to provide and display clean screenshots.
The iOS subset contains 303 screenshots from an iPhone 14 Pro at a resolution of $1179 \times 2556$.
The Android subset contains 102 screenshots from an AT\&T Maestro 3 at $720 \times 1600$ and 131 screenshots from a Samsung Galaxy S20 Ultra 5G at $1080 \times 2400$.
Each screenshot is displayed on its source smartphone, which is used consistently across the corresponding 16 physical capture settings.

Adaptive brightness is disabled on all smartphones.
Indoor captures use a fixed manual screen brightness, while outdoor captures use the maximum brightness setting.
Indoor-Dark is captured with the room lights turned off, and Indoor-Lamp uses normal laboratory lighting.
Outdoor-Shade is captured in outdoor shaded areas, while Outdoor-Sun is captured under direct sunlight.

\smallskip\noindent\textbf{Wearable image acquisition.}\quad
Table~\ref{tab:supp_capture_settings} summarizes the image acquisition methods, saved resolutions, and export processing for the four wearable devices.

\smallskip\noindent\textbf{Frame selection and quality control.}\quad
Aria Gen 1 and Aria Gen 2 continuously record RGB video, from which we extract the images corresponding to each GUI state.
We first use a sharpness score based on the variance of the Laplacian to identify candidate frames.
Annotators then manually select frames with clear screen content and complete target regions.
Quest 3 and Galaxy XR instead capture system passthrough images directly.
Captures that do not satisfy the acquisition requirements are retaken.
For all devices, observations with incorrect screen content or targets cut off by the image boundary are recaptured or replaced.

\smallskip\noindent\textbf{Image processing.}\quad
Apart from the orientation adjustment and format conversion listed in Table~\ref{tab:supp_capture_settings}, no additional cropping, resizing, image enhancement, or lens undistortion is applied during dataset preparation.
Brightness-enhanced images available in the annotation interface are used only for visual inspection and are not included in the benchmark.
Model evaluation uses the stored physical images.

\begin{table}[t]
\centering
\caption{Image acquisition and export settings for the four wearable devices.
Resolutions refer to the physical images stored in GlassGround.}
\label{tab:supp_capture_settings}
\small
\setlength{\tabcolsep}{3pt}
\begin{tabular}{llcl}
\toprule
Device & Acquisition & Resolution & Processing \\
\midrule
Aria Gen 1 & RGB video, 20 fps & $1408 \times 1408$ & $90^\circ$ CW \\
Aria Gen 2 & RGB video, 10 fps & $2560 \times 1920$ & None \\
Quest 3 & Passthrough & $3840 \times 2160$ & None \\
Galaxy XR & Passthrough & $2560 \times 2560$ & HEIC to JPEG \\
\bottomrule
\end{tabular}
\end{table}

\begin{figure}[t]
\centering
\includegraphics[width=\columnwidth]{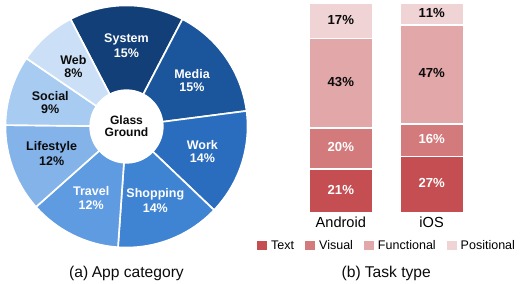}
\caption{\textbf{Benchmark composition.}
(a) Distribution of GUI states across eight application categories.
(b) Distribution of the four grounding instruction types for Android and iOS.}
\label{fig:supp_composition}
\end{figure}

\begin{figure*}[t]
\centering
\includegraphics[width=0.82\textwidth]{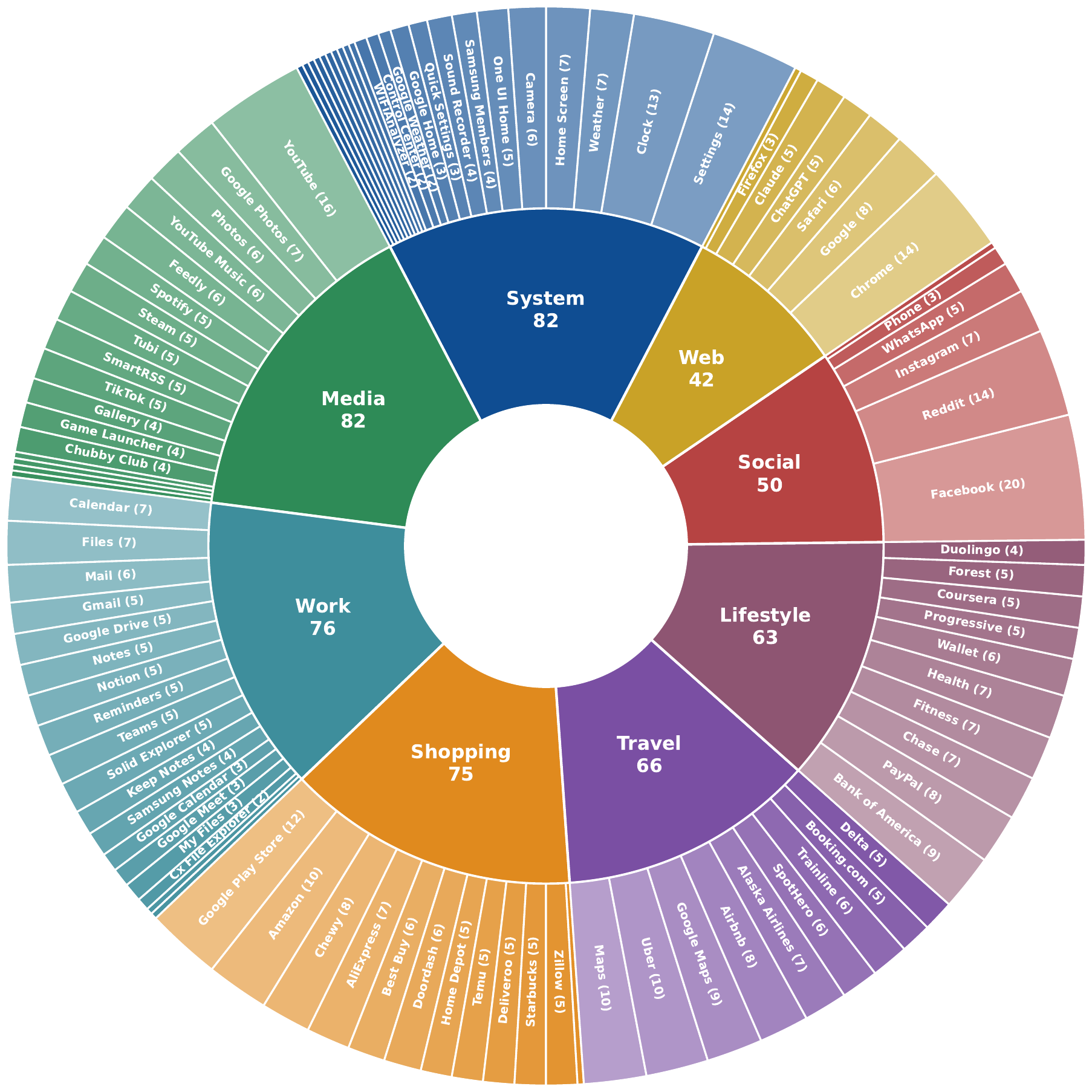}
\caption{\textbf{Application distribution in GlassGround.}
The inner ring shows the number of GUI states across eight application categories.
The outer ring breaks down each category into individual applications, with the number of GUI states indicated in parentheses.}
\label{fig:supp_taxonomy}
\end{figure*}

\subsection{Clean Side Task Annotation}
\label{supp:clean_annotation}

As described in Sec.~\ref{sec:paired_task}, we construct grounding tasks exclusively from clean screenshots without consulting their paired physical observations.
We identify candidate GUI regions using OmniParser~\cite{lu2024omniparser} and OCR, assign Set-of-Mark labels~\cite{yang2023set}, and query Claude Fable~5.1~\cite{fable51} for candidate-wise task proposals.
For suitable targets, the VLM generates grounding instructions and assigns one of four instruction types:
\emph{text}, which relies on visible text,
\emph{visual}, which relies on appearance such as icons, colors, or shapes,
\emph{functional}, which identifies the target by its function, and
\emph{positional}, which relies on location, order, or spatial relations.
The VLM does not predict target regions.

Across 510 screenshots processed by the automated pipeline, 13,586 candidate regions were submitted to the VLM.
Among these, 6,275 candidates (46.19\%) were identified as suitable grounding targets.
Annotators reviewed the generated proposals, checking target validity and clickability, instruction clarity and type, and bounding-box accuracy.
Proposals failing these checks were rejected, while valid proposals could be retained or revised.
Annotators could also activate candidates rejected by the VLM and add missing targets.

Of the 2,752 final tasks originating from positive VLM proposals, 1,538 (55.89\%) remained unchanged under our comparison criterion, while 1,214 (44.11\%) were revised.
A proposal is considered unchanged if its instruction matches after normalization, its instruction type is identical, and no bounding-box edge moves by more than two pixels.
Instruction normalization ignores capitalization, punctuation, and leading action verbs such as Tap, Click, Select, and Press.

Annotators additionally recovered 188 targets initially rejected by the VLM and manually added 183 target regions.
Another 127 tasks were manually annotated on 26 pilot screenshots.
The final benchmark contains 3,250 human-verified grounding tasks across 536 clean screenshots.
Each task is paired with 16 physical observations, yielding 52,000 physical grounding instances.

\smallskip\noindent\textbf{Benchmark composition.}\quad
GlassGround covers 102 applications across eight categories.
Figure~\ref{fig:supp_taxonomy} provides a detailed breakdown of the 536 GUI states by application category and individual application.
Figure~\ref{fig:supp_composition} summarizes the category proportions and compares the distribution of grounding instruction types between Android and iOS.

\subsection{Physical Target Transfer and Quality Control}
\label{supp:physical_annotation}

\smallskip\noindent\textbf{Automatic screen corner initialization.}\quad
We use YOLO11n-pose~\cite{yolo11} to predict the four ordered screen corners in physical images.
The detector was initially trained on 268 manually annotated pilot images not included in GlassGround and progressively updated as additional human-verified annotations became available.
The predicted corners provide an initial estimate that annotators can directly inspect and refine, avoiding the need to locate all four corners from scratch.

Since small corner localization errors can affect the alignment of transferred GUI targets, later versions incorporate an additional target projection loss alongside the standard YOLO-pose loss.
This loss penalizes discrepancies between clean target boxes projected using predicted and human-verified screen corners.
The final detector was fine-tuned from an earlier version on 5,852 images, comprising 268 pilot images, 3,526 previously annotated GlassGround images, and 2,058 AMEX camera images.
Training used an input size of 960 pixels, 30 epochs, a batch size of 32, and an initial learning rate of 0.001.

On 5,040 subsequent physical images not used to train their corresponding detector versions, the median per-image mean corner displacement relative to human-verified annotations is 5.83 pixels, or 0.89\% of the screen diagonal.

\smallskip\noindent\textbf{Human annotation protocol.}\quad
Six annotators participated in physical image geometry verification, including two experienced annotators responsible for quality control.
Annotators corrected the predicted screen corners and inspected the target regions transferred from human-verified clean screenshots.
When transferred regions remained misaligned, annotators refined the geometric mapping through the screen corners or internal grid control points.
Physical target regions were thus obtained through verified geometric transfer rather than independent target annotation.

\smallskip\noindent\textbf{Mesh refinement.}\quad
A single homography may not fully capture local geometric deviations caused by camera imaging or image processing.
We therefore provide optional mesh refinement using thin plate spline (TPS) interpolation.
For a clean screenshot point $x$, the refined mapping to physical observation $k$ is
\begin{equation}
T_k(x) = \pi(H_k \tilde{x}) + f_k(x),
\label{eq:supp_mesh_mapping}
\end{equation}
where $H_k$ is the screen homography and $f_k$ is a TPS displacement field defined in clean screenshot coordinates.

The grid is initialized from the homography projection.
Annotators adjust selected grid points to specify local residual displacements.
Only adjusted points contribute additional TPS constraints, while the four screen corners have zero residual displacement.
The tool supports adjustable grid densities, with six rows and nine columns being the most common configuration.
The median number of adjusted control points is eight.

For mesh-refined images, six points are sampled along each edge of a clean target box, including the starting vertex but excluding the ending vertex.
The resulting 24 boundary points are transformed to construct the physical target polygon.
Without mesh refinement, the target region is transferred directly using homography.

\begin{figure*}[t]
\centering
\includegraphics[width=\textwidth]{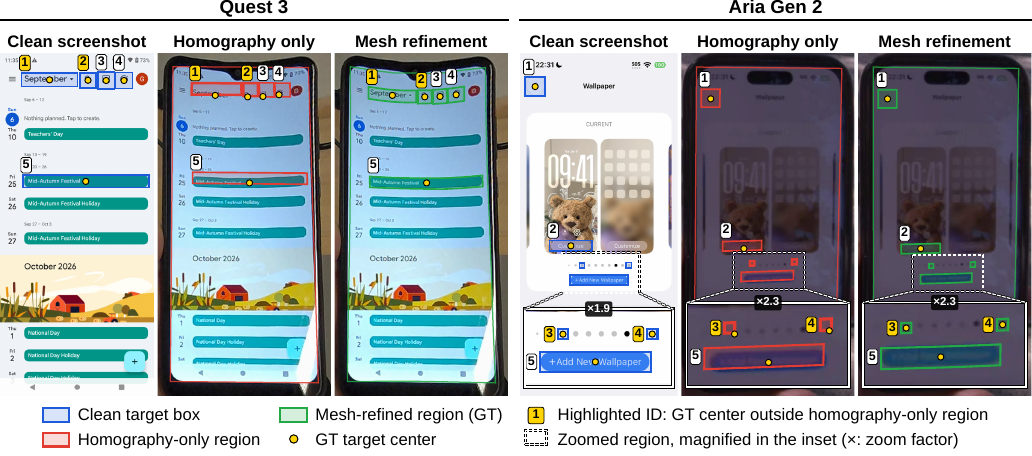}
\caption{\textbf{Effect of mesh refinement on target transfer.}
Examples from Quest 3 and Aria Gen 2.
Clean target boxes are transferred using homography alone or with mesh refinement.
The green regions represent the final transferred ground truth.
Yellow dots indicate target centers, while highlighted IDs mark centers outside the homography-only regions.
Insets enlarge small targets to illustrate local alignment differences.}
\label{fig:supp_mesh_examples}
\end{figure*}

Mesh refinement was applied to 996 physical images (11.6\%), covering 6,092 grounding instances.
Most of these images were captured by Quest 3 and Aria Gen 2.
Compared with homography alone, mesh refinement changes target centers by a median of 2.98 pixels, equivalent to 6.50\% of the target scale.
On these instances, the two mappings yield 98.04\% agreement in hit/miss verdicts.
Using homography alone produces a mean accuracy 1.10 percentage points lower across ten models.
Figure~\ref{fig:supp_mesh_examples} illustrates the resulting differences in target alignment.

\smallskip\noindent\textbf{Pilot annotation agreement.}\quad
Before large-scale annotation, two annotators independently verified the screen geometry of a designated subset without consulting each other's annotations.
The analysis covers 197 physical images and 1,166 grounding instances using homography-based target transfer.
The median per-image mean corner displacement between the two annotations is 2.25 pixels, or 0.53\% of the screen diagonal.
Table~\ref{tab:supp_annotation_reliability} shows that the resulting grounding evaluations achieve high agreement across ten models.

\begin{table}[t]
\centering
\caption{Consistency of transferred target regions under independent pilot annotation and human review.
Hit/miss verdicts are evaluated using ten Single-pass grounding models.}
\label{tab:supp_annotation_reliability}
\small
\setlength{\tabcolsep}{4pt}
\begin{tabular}{lcc}
\toprule
Metric & Pilot & Cross review \\
\midrule
Images & 197 & 8,576 \\
Instances & 1,166 & 52,000 \\
Verdict agreement (\%) & 98.98 & 99.59 \\
Cohen's $\kappa$ & 0.979 & 0.992 \\
Max. $|\Delta \mathrm{acc}|$ (pp) & 0.69 & 0.16 \\
\bottomrule
\end{tabular}
\end{table}

\begin{figure*}[t]
\centering
\includegraphics[width=\textwidth]{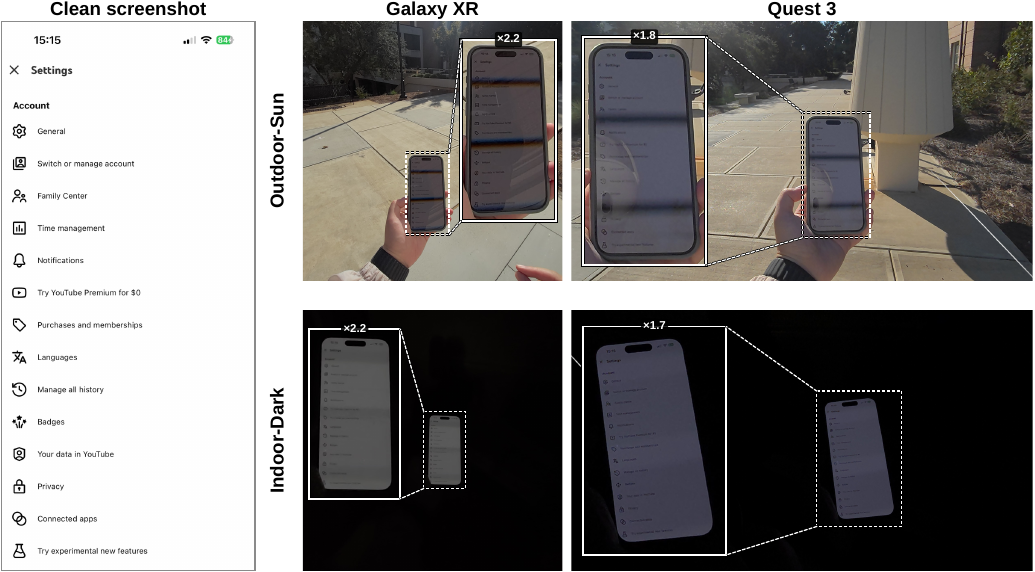}
\caption{\textbf{Device-specific imaging artifacts under different lighting conditions.}
The left panel shows the clean screenshot, while the remaining panels show the same GUI state captured by Galaxy XR and Quest 3 under Outdoor-Sun and Indoor-Dark conditions.
Outdoor-Sun captures exhibit horizontal dark bands and screen reflections.
Under Indoor-Dark, the Galaxy XR example shows reduced contrast and loss of fine text detail, while the Quest 3 example retains relatively clearer text.
Dashed boxes identify the screen regions enlarged in the insets, with magnification factors indicated.}
\label{fig:supp_degradation_examples}
\end{figure*}

\smallskip\noindent\textbf{Human cross verification.}\quad
All 8,576 physical images were annotated and reviewed by at least two annotators.
Reviewers checked the alignment of transferred target regions and refined the screen geometry when necessary.
Some images received further review by additional annotators.

We assess the effect of this process by comparing the initial human annotations with the final reviewed geometry across all 8,576 images and 52,000 grounding instances.
A target region is considered modified when its vertex count changes or any vertex coordinate differs by more than 0.5 pixels.
Review resulted in changes to 3,583 images (41.78\%) and 40.08\% of grounding instances.
The median mean corner displacement among revised images is 0.58\% of the screen diagonal, reflecting mostly fine geometric adjustments.
A smaller subset of 703 instances (1.35\%) has an IoU below 0.5 between the initial and final transferred regions.

Despite these refinements, the grounding verdict agreement remains 99.59\% across ten models.
On instances with modified target regions, agreement is 98.97\%.
Mean accuracy changes by only 0.10 percentage points, and model rankings remain unchanged.
These findings support the stability of GlassGround evaluation under human annotation revisions.

\begin{figure*}[t]
\centering
\includegraphics[width=\textwidth]{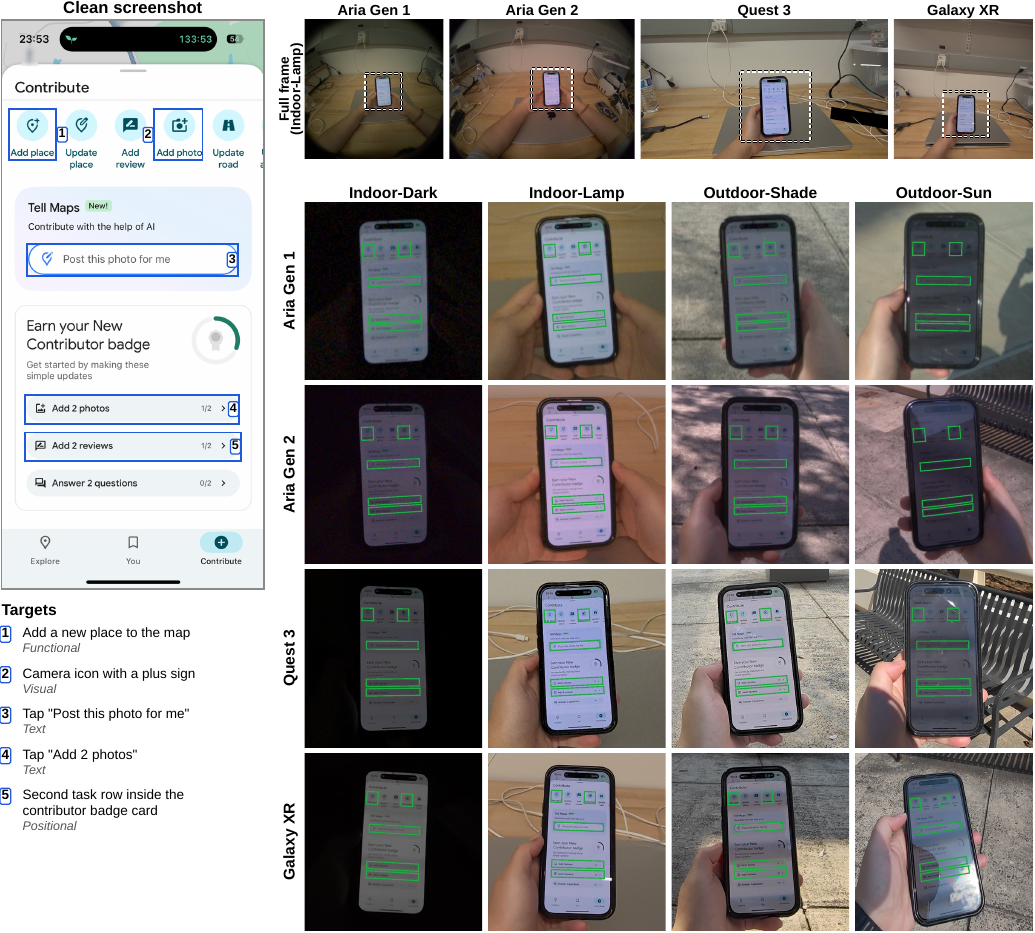}
\caption{\textbf{Complete paired observations in GlassGround.}
The left panel shows a clean screenshot with five annotated grounding targets and their instructions.
The upper-right panels show full-frame Indoor-Lamp observations from Aria Gen 1, Aria Gen 2, Quest 3, and Galaxy XR.
The lower-right grid shows cropped views of the same GUI state under all 16 combinations of four wearable devices and four lighting conditions.
Blue boxes denote clean target regions, while green polygons indicate physical ground truth transferred through human-verified geometry.
Crops and overlays are used for visualization only.}
\label{fig:supp_paired_observations}
\end{figure*}

\subsection{Device Specific Imaging Artifacts}
\label{supp:device_artifacts}

Figure~\ref{fig:supp_degradation_examples} illustrates representative imaging artifacts associated with wearable devices and lighting conditions, complementing the observations discussed in Sec.~\ref{sec:gap_magnitude}.
The examples show the same GUI state captured by Galaxy XR and Quest 3 under Outdoor-Sun and Indoor-Dark conditions.

Under Outdoor-Sun, both devices exhibit horizontal dark bands that partially obscure on-screen text and icons.
These bands are more prominent in the Galaxy XR example, while the Quest 3 capture also shows visible screen reflections.
Such banding may arise from a mismatch between display refresh or brightness modulation and camera exposure timing.
The final appearance may also be influenced by device-specific image processing.

Under Indoor-Dark, the Galaxy XR example exhibits a washed-out screen appearance with reduced local contrast and loss of fine text detail.
In comparison, the Quest 3 example preserves relatively legible menu text.
Although the overall interface structure remains visible in both observations, the readability of individual elements differs.
These examples illustrate variations in imaging artifacts rather than quantify their impact on grounding accuracy.

\subsection{Qualitative Benchmark Examples}
\label{supp:benchmark_examples}

Figure~\ref{fig:supp_paired_observations} shows a complete paired example from GlassGround.
The clean screenshot contains five grounding tasks spanning all four instruction types: functional, visual, text, and positional.
The figure includes full-frame observations from all four wearable devices under Indoor-Lamp and cropped views of all 16 device and lighting combinations.
Although the GUI state, instructions, and targets remain fixed, screen scale, position, perspective, and readability vary across observations.

\section{Evaluation Details}
\label{supp:evaluation_details}


\subsection{Model Configurations and Evaluation Protocol}
\label{supp:model_eval_setup}

\smallskip\noindent\textbf{Open-source models.}\quad
We evaluate ten open-source models locally using vLLM with bfloat16 precision and greedy decoding.
We use a temperature of 0, a maximum generation length of 512 tokens, and a random seed of 42.
Inference runs on NVIDIA RTX A6000, L40S, and A100 GPUs with tensor parallelism of 1 or 2 according to model size.
Each request contains one image and its grounding instruction.
For each model, the checkpoint, prompt template, and decoding settings remain fixed across observation conditions.
Each model uses its own visual processor with the image budget and inference mode listed in Table~\ref{tab:supp_model_config}.

\begin{table}[!t]
\centering
\scriptsize
\setlength{\tabcolsep}{2.5pt}
\caption{\textbf{Model inference configurations.}
Image budgets indicate pixel limits or native processor settings.
Models with pixel limits resize oversized inputs.
Prompt identifiers refer to the full templates provided at the end of the supplementary material.}
\label{tab:supp_model_config}
\begin{tabular}{@{}lllcc@{}}
\toprule
Model & Image budget & Inference mode & Output & Prompt \\
\midrule
MAI-UI-8B            & 4.19M px  & Native reasoning & Point & P1 \\
GUI-Owl-1.5-8B / 32B & 10.04M px & Direct           & Point & P2 \\
KV-Ground-8B         & 10.04M px & Direct           & Point & P2 \\
Indeed-UI-32B        & 10.04M px & Direct           & Point & P2 \\
UI-Venus-2-9B        & 4.80M px  & Thinking off     & Box   & P3 \\
UI-Mate-27B          & 4.80M px  & Thinking off     & Box   & P3 \\
Holo3.1-35B-A3B      & 4.19M px  & Thinking         & Point & P4 \\
Muse Glimmer-30B     & 4,096 tokens & Low reasoning  & Point & P5 \\
Qwen3.8-27B          & 4.80M px  & Thinking off     & Point & P6 \\
\midrule
Claude Sonnet 5.5    & Long side $\leq$2,576 px$^\ast$ & Adaptive, low & Point & P4 \\
GPT-6 Luna           & \texttt{detail=original} & Low reasoning & Point & P4 \\
Jev 1.14             & Long side $\leq$1,568 px & N/A & Box & P7 \\
\bottomrule
\end{tabular}
\par\vspace{2pt}
{\scriptsize $^\ast$Resized by the API.\par}
\end{table}

\smallskip\noindent\textbf{Closed-source models.}\quad
We evaluate Claude Sonnet 5.5, GPT-6 Luna, and Jev 1.14 through their respective APIs.
Claude Sonnet 5.5 uses adaptive thinking with low effort and a maximum output length of 8,192 tokens.
GPT-6 Luna uses low reasoning effort and a maximum output length of 16,000 tokens.
Both use the same grounding prompt, with images submitted according to their respective API interfaces.
Jev 1.14 uses a structured grounding interface that returns normalized bounding boxes.
All three models are evaluated on the same fixed subset described in Sec.~\ref{supp:closed_source_subset}.

\smallskip\noindent\textbf{Prediction parsing and scoring.}\quad
We convert model outputs into two-dimensional points in the original coordinate system of each observation.
Open-source models, Claude Sonnet 5.5, and GPT-6 Luna use a coordinate scale of 0 to 1000.
Jev 1.14 returns bounding boxes on a scale of 0 to 1.
For models predicting bounding boxes, we use the box center as the grounding point.
Normalized coordinates are mapped using the original image dimensions.
A prediction is correct when its point lies inside the corresponding ground-truth polygon.
Unparseable outputs or invalid final predictions are counted as incorrect.
Accuracy is the proportion of correct predictions among the evaluation instances.

\smallskip\noindent\textbf{Intermediate observation construction.}\quad
For each task, Resized uniformly scales the clean screenshot so that its target area matches the corresponding physical target area.
The scaled screenshot is placed on a gray background with the target center aligned to its position in the physical observation.
Warped projects the clean screenshot onto the screen region of the physical image using a homography from human-verified screen corners.
The surrounding background is retained from the physical observation.
For both conditions, target regions are transformed consistently with the image construction.

\smallskip\noindent\textbf{Rectified observations and scoring.}\quad
We construct Rectified observations by applying a perspective transformation to each physical image using either human-verified screen corners (Oracle) or automatically predicted corners (Auto).
Both settings use the same rectification procedure and grounding inference configuration, differing only in the source of screen corners.
The rectified screen is mapped to a front-facing canvas matching the clean screenshot dimensions.

Predicted screen corners may contain localization errors, so target positions in the rectified image do not necessarily align exactly with their original bounding boxes in the clean screenshot.
Furthermore, some physical target regions include local mesh refinement that is not captured by a single homography.
We therefore do not score predictions directly against clean target boxes in the rectified image.
Instead, we map each predicted point back to the original physical image using the inverse of the transformation applied during rectification.
The mapped point is then evaluated against the final human-verified physical ground-truth polygon.
This procedure avoids treating geometric misalignment with the clean target box as a grounding error while preserving the effect of rectification quality on model predictions.

\subsection{ReGround Implementation}
\label{supp:reground_details}

ReGround directly reuses the Single-pass prediction as its initial result.
Given a valid initial point, we extract a crop centered on the prediction, with its width and height set to half of the original image dimensions, rounded down to integers.
If the crop extends beyond the image boundary, we shift the window back inside the image while preserving its size, without padding.
For open-source models, Claude Sonnet 5.5, and GPT-6 Luna, the crop is resized to the original image dimensions.
For Jev 1.14, the crop is resized to the image dimensions used in the initial API request.
We then perform a second grounding inference using the same model, instruction, prompt template, and inference configuration.

Let $(x_0,y_0)$ denote the top-left corner of the crop, and let $(w_c,h_c)$ denote its width and height.
We convert the second prediction to normalized crop coordinates $(u,v)\in[0,1]^2$, taking the center of the predicted box for box-output models.
The final prediction in the original image coordinates is
\begin{equation}
\hat{p} = (x_0 + u w_c,\; y_0 + v h_c).
\label{eq:supp_reground_mapping}
\end{equation}

\begin{table*}[t]
\centering
\caption{
\textbf{Matched-subset grounding accuracy (\%) on GlassGround.}
All models are compared on the same stratified subset of 500 clean grounding tasks and 8,000 paired physical instances.
Single-pass results cover the observation stages and rectification with annotated (Oracle) or detected (Auto) screen corners.
ReGround results are reported separately for Resized, Warped, and Physical observations.
Adapted models report mean $\pm$ standard deviation over three training seeds.
The \textbf{best} and \underline{second-best} scores in each column are highlighted, with ties sharing the same distinction.
}
\label{tab:supp_matched_results}
\footnotesize
\setlength{\tabcolsep}{0pt}
\begin{tabular*}{\textwidth}{@{\extracolsep{\fill}} l cccc cc ccc}
\toprule
\multirow{2}{*}{\textbf{Model}}
 & \multicolumn{4}{c}{\textbf{Single-pass Accuracy} $\uparrow$}
 & \multicolumn{2}{c}{\textbf{Rectified (Single-pass)} $\uparrow$}
 & \multicolumn{3}{c}{\textbf{ReGround Accuracy} $\uparrow$} \\
\cmidrule(lr){2-5}\cmidrule(lr){6-7}\cmidrule(lr){8-10}
 & Clean & Resized & Warped & Physical
 & Oracle & Auto
 & Resized & Warped & Physical \\
\midrule

\rowcolor{baselinebg}
\multicolumn{10}{c}{\textit{Closed-source VLMs}} \\

Claude Sonnet 5.5~\cite{anthropic2026sonnet55}
 & \textbf{97.20} & \textbf{90.13}
 & \textbf{94.95} & \textbf{70.29}
 & \textbf{71.49} & \textbf{69.96}
 & \textbf{96.51} & \textbf{96.98}
 & \textbf{76.83} \\

GPT-6 Luna~\cite{openai2026gpt6}
 & 95.80 & 86.56
 & \underline{88.96} & 58.35
 & 47.53 & 46.66
 & 90.39 & 91.44
 & 61.28 \\

Jev 1.14~\cite{typesafe2026jev114}
 & 83.60 & 31.03
 & 27.14 & 16.53
 & 48.91 & 49.18
 & 49.76 & 48.28
 & 31.35 \\

\midrule
\rowcolor{baselinebg}
\multicolumn{10}{c}{\textit{Open-source baselines}} \\

MAI-UI-8B~\cite{zhou2025mai}
 & 95.60 & 84.43
 & 76.01 & 45.84
 & 62.39 & 62.65
 & 93.20 & 90.41
 & 59.26 \\

GUI-Owl-1.5-8B~\cite{xu2026mobile}
 & 94.20 & 82.38
 & 74.20 & 44.33
 & 56.33 & 56.25
 & 89.68 & 86.29
 & 54.41 \\

GUI-Owl-1.5-32B~\cite{xu2026mobile}
 & 96.00 & 83.51
 & 76.75 & 48.84
 & 61.18 & 61.25
 & 90.60 & 87.98
 & 57.03 \\

KV-Ground-8B~\cite{kvground2026}
 & 94.80 & 84.19
 & 77.65 & 48.06
 & 59.90 & 59.83
 & 90.68 & 87.11
 & 56.19 \\

UI-Venus-2-9B~\cite{team2026ui}
 & 95.60 & 87.95
 & 84.11 & 54.76
 & 65.89 & 65.81
 & 93.69 & 91.30
 & 63.89 \\

UI-Mate-27B~\cite{ding2026ui}
 & \underline{96.80} & 86.13
 & 81.68 & 50.04
 & 64.90 & 64.99
 & 93.33 & 91.51
 & 60.08 \\

Holo3.1-35B-A3B~\cite{hai2026holo31}
 & 95.80 & \underline{88.35}
 & 78.16 & 46.46
 & 63.09 & 62.93
 & 93.55 & 90.78
 & 59.89 \\

Indeed-UI-32B~\cite{indeedui2026}
 & \textbf{97.20} & 84.60
 & 78.24 & 50.31
 & 61.41 & 61.34
 & 91.29 & 88.71
 & 57.56 \\

\midrule[0.3pt]

Muse Glimmer-30B~\cite{museglimmer2026}
 & 96.00 & 85.58
 & 67.70 & 40.29
 & \underline{69.60} & 69.13
 & 94.65 & \underline{93.09}
 & 62.98 \\

Qwen3.8-27B~\cite{qwen38}
 & 92.00 & 88.30
 & 83.46 & 55.99
 & 67.03 & 67.33
 & \underline{94.74} & 92.83
 & 64.34 \\

\midrule
\textbf{Open-source Avg.}
 & 95.40 & 85.54
 & 77.80 & 48.49
 & 63.17 & 63.15
 & 92.54 & 90.00
 & 59.56 \\

\midrule
\rowcolor{oursbg}
\multicolumn{10}{c}{\textit{Adapted models (mean $\pm$ std., 3 seeds)}} \\

UI-Venus-2-9B + \textbf{PD}
 & $96.13_{\pm 0.42}$
 & $\mathbf{90.13}_{\pm 0.16}$
 & $87.96_{\pm 0.44}$
 & $\underline{64.21}_{\pm 0.63}$
 & $69.08_{\pm 0.55}$
 & $\underline{69.19}_{\pm 0.52}$
 & $94.15_{\pm 0.30}$
 & $92.85_{\pm 0.29}$
 & $\underline{69.57}_{\pm 0.36}$ \\

MAI-UI-8B + \textbf{PD}
 & $95.47_{\pm 0.31}$
 & $85.11_{\pm 1.01}$
 & $81.98_{\pm 0.37}$
 & $57.71_{\pm 0.32}$
 & $65.10_{\pm 0.89}$
 & $65.02_{\pm 0.58}$
 & $92.06_{\pm 0.64}$
 & $90.28_{\pm 0.72}$
 & $64.35_{\pm 0.17}$ \\

\bottomrule
\end{tabular*}
\end{table*}

If the second prediction is invalid, we retain the initial prediction.
If the initial prediction is invalid, we skip the second inference and count the instance as incorrect.
Every valid initial prediction triggers one refinement inference.
Each instance requires at most two model inferences in total, with the first prediction reused from Single-pass evaluation.
Neither ground-truth target annotations nor screen corner annotations are used during cropping or refinement.

\subsection{Evaluation Subset and Matched Comparison}
\label{supp:closed_source_subset}

\smallskip\noindent\textbf{Subset construction.}\quad
To control the inference cost of closed-source models, we construct a fixed evaluation subset of 500 grounding tasks from the 3,250 clean tasks in GlassGround.
We stratify the tasks along three dimensions: operating system (Android or iOS), instruction type (text, visual, functional, or positional), and tertiles of clean target bounding-box area.
This produces 24 strata, from which we sample tasks approximately in proportion to their frequencies in the full benchmark.
We generate candidate subsets using 50 random seeds from 0 to 49 and select seed 39, which minimizes the mean absolute accuracy difference between the full benchmark and the corresponding subset across the ten open-source models.

\smallskip\noindent\textbf{Subset size and evaluation coverage.}\quad
The selected subset contains 500 grounding tasks from 332 distinct clean GUI states.
Each task is paired with all 16 combinations of four wearable devices and four lighting conditions, yielding 8,000 physical grounding instances across 5,312 distinct physical images.
All three closed-source models are evaluated on this fixed subset under Clean, Resized, Warped, Physical, Oracle, and Auto conditions, as well as ReGround on Resized, Warped, and Physical observations.
Clean contains 500 evaluation instances, while each of the other conditions contains 8,000 instances.

\smallskip\noindent\textbf{Matched-subset comparison.}\quad
The open-source baseline results in the main paper are evaluated on the full benchmark.
For direct comparison with closed-source models, we extract the corresponding predictions from the full-benchmark open-source evaluations and recompute accuracy on the same 500-task subset.
We also evaluate the two PairDistill-adapted models on this subset and report their mean accuracy and standard deviation across three training seeds.
Table~\ref{tab:supp_matched_results} presents the resulting matched comparisons across all evaluation conditions.

\subsection{Held-Out Automatic Rectification}
\label{supp:auto_rectification}

\smallskip\noindent\textbf{Held-out evaluation protocol.}\quad
As described in Sec.~\ref{supp:physical_annotation}, the screen corner detector was trained using a subset of GlassGround physical images.
To assess its performance on unseen observations, we construct a held-out evaluation subset that does not overlap with the detector's training data.
This subset contains 315 distinct clean GUI states and 1,883 grounding tasks, yielding 30,128 physical grounding instances.
Neither the physical images nor their corresponding clean GUI states were used to train the corner detector.
We evaluate Physical, Oracle, and Auto on identical held-out instances using the same model configurations and the scoring protocol described in Sec.~\ref{supp:model_eval_setup}.

\smallskip\noindent\textbf{Downstream grounding results.}\quad
Table~\ref{tab:supp_heldout_rectification} reports the results for ten open-source models and two PairDistill-adapted models.
The mean Physical Single-pass accuracy across the ten open-source models is 50.01\%.
Oracle rectification using human-verified screen corners increases this accuracy to 64.90\%, while Auto rectification using predicted corners achieves 64.73\%.
The difference between Oracle and Auto is less than 0.2 percentage points.
The absolute accuracy difference between the two settings is at most 0.4 percentage points for every open-source baseline.

The two PairDistill-adapted models exhibit a similar pattern.
For UI-Venus-2-9B with PairDistill, Oracle and Auto achieve 71.20\% and 71.10\% accuracy, respectively.
For MAI-UI-8B with PairDistill, the corresponding accuracies are 66.87\% and 66.71\%.
These results show that automatically predicted screen corners support effective rectification on physical observations unseen during detector training, achieving downstream grounding accuracy close to that obtained with human-verified corners.

\begin{table}[t]
\centering
\caption{
\textbf{Grounding accuracy (\%) with automatic rectification on held-out physical observations.}
Results are evaluated on 30,128 physical grounding instances from 315 clean GUI states not used for training the corner detector.
Oracle uses human-verified screen corners, while Auto uses predicted corners.
All predictions are scored using the protocol in Sec.~\ref{supp:model_eval_setup}.
PairDistill (PD) results are reported as mean $\pm$ standard deviation over three training seeds.
Model abbreviations follow Table~\ref{tab:supp_model_config}.
This subset differs from the full benchmark, so absolute accuracies should not be directly compared across the two evaluation sets.
}
\label{tab:supp_heldout_rectification}
\scriptsize
\setlength{\tabcolsep}{2pt}
\begin{tabular*}{\columnwidth}{@{\extracolsep{\fill}}lccc@{}}
\toprule
Model & Physical & Oracle & Auto \\
\midrule
MAI-UI-8B & 46.85 & 64.26 & 64.31 \\
Owl-8B & 46.15 & 58.20 & 58.20 \\
Owl-32B & 51.13 & 62.84 & 62.68 \\
KV-Ground & 49.73 & 61.83 & 61.86 \\
UI-Venus-2-9B & 56.69 & 67.72 & 67.42 \\
UI-Mate-27B & 51.29 & 66.25 & 66.06 \\
Holo3.1 & 48.07 & 65.33 & 65.33 \\
Indeed-UI & 52.48 & 62.92 & 62.59 \\
Muse Glimmer & 40.40 & 70.46 & 70.08 \\
Qwen3.8 & 57.29 & 69.14 & 68.79 \\
\midrule
\textbf{Open-source Avg.} & 50.01 & 64.90 & 64.73 \\
\midrule
Venus + \textbf{PD}
 & $65.63_{\pm 0.58}$
 & $71.20_{\pm 0.56}$
 & $71.10_{\pm 0.55}$ \\
MAI + \textbf{PD}
 & $59.52_{\pm 0.28}$
 & $66.87_{\pm 0.48}$
 & $66.71_{\pm 0.45}$ \\
\bottomrule
\end{tabular*}
\end{table}

\section{Additional Analysis Results}
\label{supp:additional_analysis}


\subsection{Condition-Wise Performance of Open-Source Models}
\label{supp:condition_open}

Table~\ref{tab:supp_full_conditions} reports Physical Single-pass accuracy across all 16 device and lighting combinations for ten open-source baselines and the SFT and PairDistill variants of two backbone models.
Each condition contains 3,250 grounding instances.
Performance varies substantially across wearable devices and lighting conditions.
Outdoor-Sun yields the lowest accuracy among the four lighting conditions for every open-source baseline on each device.

PairDistill improves both backbone models across the evaluated device and lighting combinations, including conditions not represented in its training data.
For Aria Gen 1, the two indoor conditions are seen during PairDistill training, whereas Outdoor-Shade and Outdoor-Sun are unseen.
On these two unseen conditions, PairDistill improves UI-Venus-2-9B over its base model by approximately 14.9 and 8.5 percentage points, respectively.
The corresponding improvements for MAI-UI-8B are approximately 10.4 and 6.1 percentage points.
Both PairDistill variants also outperform their corresponding SFT configurations on these conditions.
Although absolute accuracy remains limited under these challenging observations, the results show that the improvements extend beyond the device and lighting combinations encountered during training.

\begin{table*}[t]
\centering
\caption{
\textbf{Physical Single-pass accuracy (\%) across 16 device--lighting conditions on the full GlassGround benchmark.}
Each condition contains 3,250 grounding instances.
Panel (a) reports ten open-source baselines and their mean.
Panel (b) reports SFT and PairDistill (PD) variants of UI-Venus-2-9B and MAI-UI-8B as mean $\pm$ standard deviation over three training seeds.
$^\ast$Seen device--lighting combination in PairDistill training data.
Model abbreviations correspond to the full model names in Table~\ref{tab:supp_model_config}.
}
\label{tab:supp_full_conditions}
\scriptsize
\setlength{\tabcolsep}{1.6pt}
\begin{tabular*}{\textwidth}{@{\extracolsep{\fill}} ll *{11}{r} @{}}
\toprule
\rowcolor{baselinebg}
\multicolumn{13}{l}{\textit{(a) Open-source baselines}} \\
Device & Lighting & MAI & Owl-8B & Owl-32B & KV & Indeed & Venus & Mate & Holo & Muse & Qwen & Avg. \\
\midrule
Aria Gen 1 & Dark$^\ast$ & 20.68 & 13.94 & 21.69 & 19.14 & 22.18 & 29.32 & 21.51 & 19.29 & 3.51 & 29.35 & 20.06 \\
 & Lamp$^\ast$ & 22.65 & 15.23 & 20.86 & 19.17 & 22.00 & 30.03 & 26.95 & 17.94 & 1.51 & 30.40 & 20.67 \\
 & Shade & 11.23 & 5.78 & 10.15 & 7.51 & 10.71 & 11.69 & 8.89 & 10.00 & 0.46 & 13.05 & 8.95 \\
 & Sun & 7.35 & 3.75 & 7.14 & 4.71 & 7.20 & 7.17 & 5.05 & 6.49 & 0.12 & 7.26 & 5.62 \\
\addlinespace[2pt]
Aria Gen 2 & Dark & 49.54 & 45.42 & 53.82 & 50.65 & 55.14 & 65.63 & 57.54 & 55.51 & 49.32 & 66.71 & 54.93 \\
 & Lamp & 59.60 & 63.26 & 68.40 & 67.45 & 70.06 & 77.20 & 72.15 & 65.08 & 36.62 & 78.65 & 65.85 \\
 & Shade & 38.43 & 33.75 & 44.18 & 39.45 & 45.88 & 53.14 & 42.58 & 27.63 & 13.91 & 54.89 & 39.38 \\
 & Sun & 19.11 & 13.63 & 21.02 & 16.58 & 21.82 & 26.09 & 16.80 & 11.78 & 3.75 & 26.09 & 17.67 \\
\addlinespace[2pt]
Quest 3 & Dark$^\ast$ & 76.22 & 76.92 & 74.28 & 79.45 & 75.91 & 83.17 & 78.98 & 79.94 & 82.06 & 84.25 & 79.12 \\
 & Lamp & 78.92 & 80.58 & 79.66 & 82.37 & 81.05 & 86.28 & 82.62 & 83.82 & 82.86 & 86.95 & 82.51 \\
 & Shade & 73.94 & 74.46 & 75.72 & 76.28 & 77.26 & 81.97 & 77.57 & 77.17 & 76.92 & 82.22 & 77.35 \\
 & Sun$^\ast$ & 45.60 & 43.75 & 47.82 & 47.35 & 47.94 & 53.57 & 45.11 & 45.38 & 40.40 & 52.77 & 46.97 \\
\addlinespace[2pt]
Galaxy XR & Dark & 59.20 & 56.92 & 63.66 & 62.18 & 64.55 & 68.25 & 64.40 & 63.35 & 65.60 & 71.45 & 63.96 \\
 & Lamp & 72.43 & 75.29 & 79.78 & 79.11 & 81.66 & 84.09 & 79.97 & 79.69 & 76.68 & 85.20 & 79.39 \\
 & Shade & 64.09 & 70.95 & 77.45 & 76.49 & 80.00 & 80.83 & 77.66 & 70.77 & 68.18 & 82.71 & 74.91 \\
 & Sun & 33.78 & 34.74 & 39.54 & 38.95 & 41.14 & 43.57 & 38.46 & 34.62 & 34.18 & 44.68 & 38.37 \\
\midrule
Overall & & 45.80 & 44.28 & 49.07 & 47.93 & 50.28 & 55.13 & 49.77 & 46.78 & 39.76 & 56.04 & 48.48 \\
\bottomrule
\end{tabular*}

\par\vspace{7pt}

\setlength{\tabcolsep}{4pt}
\begin{tabular*}{\textwidth}{@{\extracolsep{\fill}} ll cccc @{}}
\toprule
\rowcolor{oursbg}
\multicolumn{6}{l}{\textit{(b) Adapted models (mean $\pm$ std., 3 training seeds)}} \\
Device & Lighting & Venus+SFT & Venus+PD & MAI+SFT & MAI+PD \\
\midrule
Aria Gen 1 & Dark$^\ast$ & $40.82_{\pm 1.27}$ & $44.50_{\pm 1.36}$ & $31.12_{\pm 3.03}$ & $35.88_{\pm 0.67}$ \\
 & Lamp$^\ast$ & $46.45_{\pm 0.43}$ & $51.26_{\pm 0.96}$ & $35.10_{\pm 4.48}$ & $42.76_{\pm 1.46}$ \\
 & Shade & $22.98_{\pm 0.94}$ & $26.62_{\pm 1.04}$ & $18.40_{\pm 2.05}$ & $21.65_{\pm 0.30}$ \\
 & Sun & $13.77_{\pm 0.96}$ & $15.70_{\pm 0.50}$ & $12.02_{\pm 1.02}$ & $13.47_{\pm 0.05}$ \\
\addlinespace[2pt]
Aria Gen 2 & Dark & $70.23_{\pm 1.02}$ & $72.91_{\pm 0.67}$ & $59.34_{\pm 3.12}$ & $63.59_{\pm 0.43}$ \\
 & Lamp & $83.30_{\pm 0.40}$ & $85.23_{\pm 0.53}$ & $71.87_{\pm 2.45}$ & $75.94_{\pm 0.22}$ \\
 & Shade & $62.24_{\pm 0.23}$ & $65.86_{\pm 1.07}$ & $50.64_{\pm 1.73}$ & $55.45_{\pm 0.87}$ \\
 & Sun & $33.99_{\pm 1.44}$ & $39.27_{\pm 0.42}$ & $26.16_{\pm 1.75}$ & $31.52_{\pm 0.97}$ \\
\addlinespace[2pt]
Quest 3 & Dark$^\ast$ & $85.20_{\pm 0.37}$ & $85.99_{\pm 0.30}$ & $79.60_{\pm 0.87}$ & $81.73_{\pm 0.57}$ \\
 & Lamp & $88.25_{\pm 0.12}$ & $89.08_{\pm 0.18}$ & $83.64_{\pm 1.09}$ & $84.86_{\pm 0.23}$ \\
 & Shade & $84.66_{\pm 0.17}$ & $85.77_{\pm 0.87}$ & $79.06_{\pm 1.59}$ & $81.66_{\pm 0.28}$ \\
 & Sun$^\ast$ & $59.55_{\pm 0.60}$ & $64.61_{\pm 1.37}$ & $53.76_{\pm 1.70}$ & $57.88_{\pm 0.44}$ \\
\addlinespace[2pt]
Galaxy XR & Dark & $74.11_{\pm 0.76}$ & $75.67_{\pm 0.42}$ & $67.35_{\pm 2.24}$ & $69.28_{\pm 0.17}$ \\
 & Lamp & $88.06_{\pm 0.59}$ & $89.04_{\pm 0.40}$ & $81.24_{\pm 1.75}$ & $83.91_{\pm 0.29}$ \\
 & Shade & $85.84_{\pm 0.26}$ & $87.21_{\pm 0.11}$ & $76.52_{\pm 1.36}$ & $81.18_{\pm 0.63}$ \\
 & Sun & $49.84_{\pm 0.89}$ & $53.01_{\pm 0.75}$ & $42.98_{\pm 1.48}$ & $46.68_{\pm 0.37}$ \\
\midrule
Overall & & $61.83_{\pm 0.39}$ & $64.48_{\pm 0.53}$ & $54.30_{\pm 1.84}$ & $57.96_{\pm 0.21}$ \\
\bottomrule
\end{tabular*}
\end{table*}

\subsection{Condition-Wise Performance of Closed-Source Models}
\label{supp:condition_closed}

Table~\ref{tab:supp_closed_conditions} compares three closed-source models, the mean of ten open-source baselines, and two PairDistill-adapted models on the same fixed 500-task subset.
All condition-wise results are evaluated on identical tasks and physical observations.

Claude Sonnet 5.5 achieves 59.20\% and 66.40\% accuracy on Aria Gen 1 under Indoor-Dark and Indoor-Lamp, respectively, substantially exceeding the corresponding open-source averages of 20.24\% and 20.56\%.
These results show that the physical observations in these two conditions retain visual information sufficient for successful grounding on a substantial fraction of tasks.
Thus, the low accuracy of open-source baselines cannot be attributed solely to missing visual information.

Across the entire matched subset, Claude Sonnet 5.5 achieves the highest Physical accuracy at 70.29\%, compared with 64.21\% for UI-Venus with PairDistill.
Nevertheless, PairDistill substantially narrows the gap between the open-source backbone and the stronger closed-source model.
On the unseen Aria Gen 1 Outdoor-Shade and Outdoor-Sun conditions, UI-Venus with PairDistill obtains 27.13\% and 15.60\% accuracy, compared with 26.60\% and 13.60\% for Claude Sonnet 5.5.
These are similar observed accuracies on identical evaluation instances, without establishing statistical equivalence.
On another unseen condition, Aria Gen 2 under Outdoor-Sun, UI-Venus with PairDistill achieves 38.60\%, compared with 33.20\% for Claude Sonnet 5.5.
Together with the overall results, these comparisons illustrate the potential of PairDistill to improve grounding under unseen physical capture conditions while showing that a performance gap to stronger closed-source models remains.

\begin{table*}[t]
\centering
\caption{
\textbf{Physical Single-pass accuracy (\%) on the matched 500-task subset.}
All models are evaluated on the same tasks and physical observations.
Each device--lighting condition contains 500 grounding instances.
The open-source average is computed across ten baseline models.
PairDistill (PD) results are reported as mean $\pm$ standard deviation over three training seeds.
$^\ast$Seen device--lighting combination in PairDistill training data.
}
\label{tab:supp_closed_conditions}
\scriptsize
\setlength{\tabcolsep}{4pt}
\begin{tabular*}{\textwidth}{@{\extracolsep{\fill}} ll rrrrcc @{}}
\toprule
Device & Lighting & Sonnet 5.5 & GPT-6 Luna & Jev 1.14 & Open Avg. & Venus+PD & MAI+PD \\
\midrule
Aria Gen 1 & Dark$^\ast$ & 59.20 & 26.00 & 7.80 & 20.24 & $45.27_{\pm 2.10}$ & $37.00_{\pm 2.16}$ \\
 & Lamp$^\ast$ & 66.40 & 34.60 & 9.20 & 20.56 & $51.53_{\pm 1.70}$ & $44.20_{\pm 0.60}$ \\
 & Shade & 26.60 & 22.20 & 4.00 & 9.00 & $27.13_{\pm 1.51}$ & $22.80_{\pm 0.72}$ \\
 & Sun & 13.60 & 13.20 & 3.60 & 6.16 & $15.60_{\pm 1.40}$ & $13.47_{\pm 1.72}$ \\
\addlinespace[2pt]
Aria Gen 2 & Dark & 86.40 & 65.40 & 14.60 & 54.54 & $73.33_{\pm 1.33}$ & $62.67_{\pm 0.46}$ \\
 & Lamp & 90.00 & 88.80 & 15.40 & 65.70 & $83.67_{\pm 0.31}$ & $75.27_{\pm 1.01}$ \\
 & Shade & 69.60 & 62.00 & 8.00 & 39.70 & $65.33_{\pm 0.64}$ & $54.73_{\pm 2.21}$ \\
 & Sun & 33.20 & 28.00 & 5.20 & 16.90 & $38.60_{\pm 0.72}$ & $31.20_{\pm 1.22}$ \\
\addlinespace[2pt]
Quest 3 & Dark$^\ast$ & 93.80 & 81.60 & 29.40 & 79.78 & $85.33_{\pm 1.14}$ & $81.47_{\pm 0.50}$ \\
 & Lamp & 95.20 & 89.60 & 34.20 & 82.80 & $88.53_{\pm 0.12}$ & $84.33_{\pm 0.50}$ \\
 & Shade & 93.00 & 82.60 & 33.00 & 78.96 & $86.53_{\pm 1.03}$ & $82.60_{\pm 0.69}$ \\
 & Sun$^\ast$ & 69.20 & 47.60 & 18.20 & 46.22 & $62.13_{\pm 1.53}$ & $53.60_{\pm 0.72}$ \\
\addlinespace[2pt]
Galaxy XR & Dark & 85.80 & 72.00 & 19.40 & 62.74 & $76.60_{\pm 0.40}$ & $71.00_{\pm 0.69}$ \\
 & Lamp & 95.20 & 82.80 & 25.20 & 79.86 & $88.93_{\pm 0.12}$ & $83.00_{\pm 1.00}$ \\
 & Shade & 93.40 & 88.60 & 26.80 & 74.82 & $87.33_{\pm 0.23}$ & $79.80_{\pm 0.87}$ \\
 & Sun & 54.00 & 48.60 & 10.40 & 37.88 & $51.53_{\pm 0.31}$ & $46.27_{\pm 0.76}$ \\
\midrule
Overall & & 70.29 & 58.35 & 16.53 & 48.49 & $64.21_{\pm 0.63}$ & $57.71_{\pm 0.32}$ \\
\bottomrule
\end{tabular*}
\end{table*}

\subsection{Statistical Reliability of Main Findings}
\label{supp:statistical_reliability}

\smallskip\noindent\textbf{Bootstrap protocol.}\quad
A clean GUI state in GlassGround may contain multiple grounding tasks, each paired with 16 physical observations.
Consequently, evaluation instances are not fully independent.
We assess the uncertainty of the main accuracy differences using a paired cluster bootstrap at the clean GUI-state level.
For each bootstrap replicate, we sample 536 clean GUI states with replacement and retain all tasks and physical observations associated with each sampled state.
We perform 10,000 bootstrap replicates and report percentile 95\% confidence intervals.
For adapted models with three training seeds, we first average the per-instance correctness indicators across seeds before resampling.
The resulting intervals reflect uncertainty due to the composition of evaluation samples, while variation across training seeds is reported separately through the standard deviations in the main results.

\smallskip\noindent\textbf{Accuracy drops across observation stages.}\quad
As shown in Table~\ref{tab:supp_bootstrap}, the mean accuracy drops across ten open-source models are 9.39, 7.69, and 29.24 percentage points for Clean to Resized, Resized to Warped, and Warped to Physical, respectively.
Their corresponding 95\% confidence intervals are [8.74, 10.04], [7.42, 7.94], and [28.66, 29.84].
To compare the stage drops directly, we compute paired differences between them within the same bootstrap replicates.
The Warped-to-Physical drop exceeds the other two drops by 19.86 and 21.56 percentage points, respectively.
Both paired-difference confidence intervals have lower bounds above zero.
Thus, Warped to Physical remains the largest accuracy drop after accounting for sampling uncertainty at the GUI-state level.

\smallskip\noindent\textbf{PairDistill improvements.}\quad
Under Physical Single-pass evaluation, PairDistill improves accuracy over the original UI-Venus-2-9B and MAI-UI-8B backbones by 9.36 and 12.17 percentage points, respectively.
The corresponding 95\% confidence intervals are [8.81, 9.89] and [11.54, 12.80].
Relative to SFT configurations that also use scale augmentation, PairDistill further improves accuracy by 2.65 and 3.66 percentage points, with intervals of [2.39, 2.92] and [3.38, 3.96].
The lower bounds of all four intervals exceed zero, indicating positive accuracy differences after accounting for uncertainty in the evaluated GUI-state sample.

\begin{table}[t]
\centering
\caption{
\textbf{GUI-state-level paired bootstrap confidence intervals.}
Point estimates and percentile 95\% confidence intervals are reported in percentage points.
We use 10,000 cluster resamples of 536 clean GUI states.
All adaptation comparisons use Physical Single-pass accuracy.
Both SFT and PairDistill use scale augmentation.
For adapted models, correctness is averaged across three training seeds at the instance level before resampling.
}
\label{tab:supp_bootstrap}
\footnotesize
\setlength{\tabcolsep}{3pt}
\begin{tabular*}{\columnwidth}{@{\extracolsep{\fill}}lrr@{}}
\toprule
Comparison & Difference & 95\% CI \\
\midrule
Clean $\to$ Resized drop
 & 9.39 & [8.74, 10.04] \\
Resized $\to$ Warped drop
 & 7.69 & [7.42, 7.94] \\
Warped $\to$ Physical drop
 & 29.24 & [28.66, 29.84] \\
\addlinespace[2pt]
$(W\to P)-(C\to R)$
 & 19.86 & [18.99, 20.72] \\
$(W\to P)-(R\to W)$
 & 21.56 & [20.96, 22.16] \\
\midrule
Venus: Base $\to$ PD
 & 9.36 & [8.81, 9.89] \\
MAI: Base $\to$ PD
 & 12.17 & [11.54, 12.80] \\
Venus: SFT $\to$ PD
 & 2.65 & [2.39, 2.92] \\
MAI: SFT $\to$ PD
 & 3.66 & [3.38, 3.96] \\
\bottomrule
\end{tabular*}
\end{table}



\section{PairDistill Details}
\label{supp:pairdistill_details}

\subsection{Paired Training Data Collection}
\label{supp:pairdistill_training_data}

\begin{figure}[t]
\centering
\includegraphics[width=\linewidth]{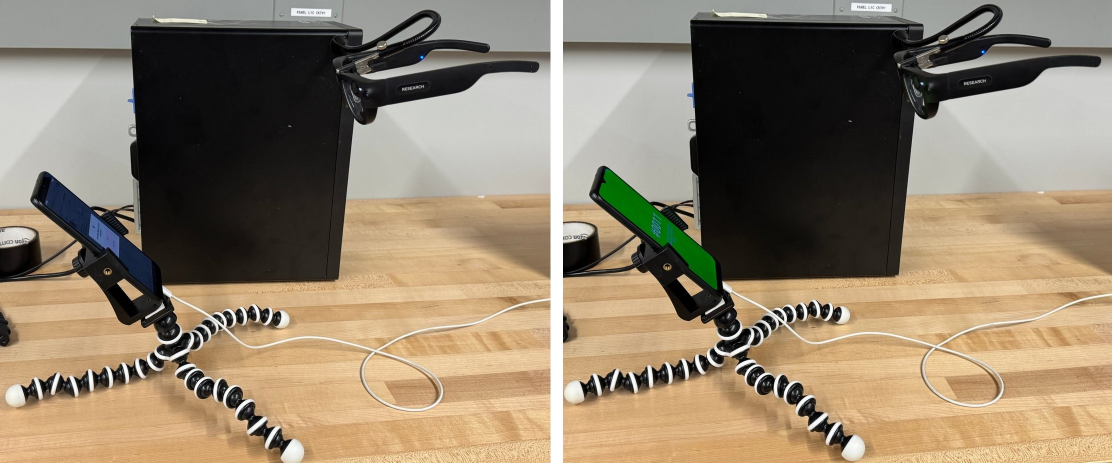}
\caption{\textbf{Training data capture setup (Aria Gen 1).} The smartphone and the wearable device are mounted on fixed holders. Left: an AMEX screenshot displayed in full screen. Right: a numbered green separator frame displayed between consecutive screenshots, used to segment the recorded video.}
\label{fig:supp_rig}
\end{figure}

\smallskip\noindent\textbf{Data source and capture.}\quad
We construct PairDistill training data from AMEX~\cite{chai2025amex}, which provides element functionality descriptions for approximately 21K of its more than 104K screenshots.
We use these descriptions as grounding instructions and the corresponding element bounding boxes as target regions.
Due to capture costs, we select approximately the first quarter of these annotated screenshots in filename order.
We display 5,098 distinct screenshots on a smartphone and record them with wearable devices, both mounted on fixed stands.
Their relative positions remain fixed within each capture session.
The screenshots are shown sequentially in full-screen mode, with numbered green separator frames inserted between consecutive screenshots.
Figure~\ref{fig:supp_rig} illustrates the capture setup and the separator frames used for video segmentation.
The wearable device continuously records video, which we segment using these separator frames to identify the corresponding screenshot intervals and extract physical images.
Each screenshot is captured only once under one of the four device and lighting combinations.
After human verification of screen corners, we transfer clean target boxes to the physical images using homography.
We exclude captures requiring non-homographic alignment and invalid capture records, yielding 4,880 valid pairs of clean screenshots and physical images.

\smallskip\noindent\textbf{Training and validation split.}\quad
We randomly reserve approximately 5\% of the valid image pairs for validation at the clean screenshot level using seed 42, ensuring that identical screenshots do not appear in both training and validation.
This produces 4,636 training pairs and an initial validation set of 244 pairs.
To reduce near-duplicate interfaces in validation, we resize clean screenshots to $32\times32$ grayscale thumbnails and compute normalized cross-correlation between each validation screenshot and all training screenshots.
We exclude validation samples whose maximum correlation with any training screenshot is at least 0.85.
The resulting validation set contains 83 image pairs and 1,266 grounding targets and is used for checkpoint selection.
The training set contains 4,636 image pairs and 76,224 grounding targets.
Of these, 1,960 pairs with 33,755 targets were captured using Aria Gen 1 under Indoor-Lamp, and 953 pairs with 14,782 targets under Indoor-Dark.
The remaining pairs were captured using Quest 3, comprising 945 pairs with 15,310 targets under Indoor-Dark and 778 pairs with 12,377 targets under Outdoor-Sun.

\subsection{Training Implementation}
\label{supp:pairdistill_training}

\smallskip\noindent\textbf{LoRA configuration.}\quad
We adapt UI-Venus-2-9B and MAI-UI-8B using LoRA with rank 16, scaling parameter 32, and dropout 0.05.
For MAI-UI-8B, LoRA is applied to the attention and MLP projections of all 36 language model layers.
For UI-Venus-2-9B, LoRA covers the attention projections of eight standard attention layers and 24 linear attention layers, together with the MLP projections of all 32 layers.
For both backbones, LoRA also covers the attention and feed-forward projections of the 27 vision encoder blocks, the patch embedding projection, and the patch merger.
MAI-UI-8B additionally adapts three DeepStack mergers.
This yields 52.5 million trainable parameters across 369 modules for MAI-UI-8B and 51.3 million trainable parameters across 359 modules for UI-Venus-2-9B.

\smallskip\noindent\textbf{Optimization.}\quad
We use AdamW with $\beta_1=0.9$, $\beta_2=0.999$, and weight decay 0.01.
The learning rate is $10^{-4}$ with cosine decay and linear warmup over the first 3\% of optimization steps.
Each training sample contains one physical image and one of its grounding targets.
Targets are grouped into four tiers by width relative to the image width.
In each epoch, the images are shuffled and each image contributes one target, drawn at random with a fixed seed from the smallest tier present in that image.
We use a microbatch size of one and gradient accumulation over four steps, giving an effective batch size of four.
Training runs for three epochs, with 1,159 optimizer updates per epoch and 3,477 updates in total.
The frozen base weights use bfloat16 precision, while LoRA parameters are maintained in float32 without automatic mixed-precision casting.
We enable gradient checkpointing and clip gradient norms at 1.0.

\smallskip\noindent\textbf{Scale augmentation.}\quad
Scale augmentation is applied independently to each physical training image, while the corresponding clean screenshot remains unchanged.
For each sample, we leave the image unchanged with probability 0.2, enlarge it with probability 0.4, or shrink it with probability 0.4.
For enlargement, we sample a scale factor log-uniformly from $[1,\min(s_{\max},3)]$, where $s_{\max}$ is the largest factor that permits a crop containing both the screen and target regions.
We randomly position the crop and resize it to the original image dimensions.
For shrinking, we sample a scale factor log-uniformly from $[1/3,1]$, resize the entire image, and place it at a random position on a canvas of the original dimensions.
The canvas background is filled with either a random gray value or the mean color of the original image.
We apply the same scaling and translation to the physical target quadrilateral, coordinate labels, and screen corners.
Student-side feature pooling uses the transformed target region, while teacher-side pooling uses the original target region in the clean screenshot.

\smallskip\noindent\textbf{Feature alignment.}\quad
We extract 4,096-dimensional visual features from the patch merger outputs of the frozen teacher and adapted student.
For MAI-UI-8B, the features are taken from the main patch merger rather than the DeepStack mergers.
Following Sec.~\ref{sec:paired_adaptation}, we subtract the mean visual token feature of each image before pooling token features according to their overlap with the target region.
The teacher is the frozen original backbone without LoRA.
Teacher target features are precomputed from clean screenshots and cached for training.
The visual input pixel budgets are 4,194,304 for MAI-UI-8B and 4,800,000 for UI-Venus-2-9B.
Student features are computed from physical observations during training.
We use the alignment loss $1-\cos(\mathbf{z}_p,\mathbf{z}_c)$ with weight $\lambda=1$.

\smallskip\noindent\textbf{Checkpoint selection and weight interpolation.}\quad
We evaluate checkpoints every 400 optimizer updates using the 83 AMEX validation image pairs and all 1,266 corresponding grounding targets described in Sec.~\ref{supp:pairdistill_training_data}.
Validation accuracy is measured by checking whether each predicted point falls inside the projected target quadrilateral in the physical image.
For MAI-UI-8B, we use the directly predicted point, while for UI-Venus-2-9B, we use the center of the predicted bounding box.
Unparseable predictions are counted as incorrect, with a fixed denominator of 1,266.
We use greedy decoding with maximum generation lengths of 48 tokens for MAI-UI-8B and 32 tokens for UI-Venus-2-9B.
For each training seed, we select the checkpoint with the highest validation accuracy.
Following WiSE-FT~\cite{wortsman2022robust}, we use a weight interpolation coefficient of $\alpha=0.5$ for the final adapted models.
In our LoRA implementation, this is achieved by reducing the LoRA scaling parameter from 32 during training to 16 at inference.

\section{Additional Adaptation Results}
\label{supp:additional_adaptation}

\begin{figure}[t]
\centering
\includegraphics[width=\columnwidth]{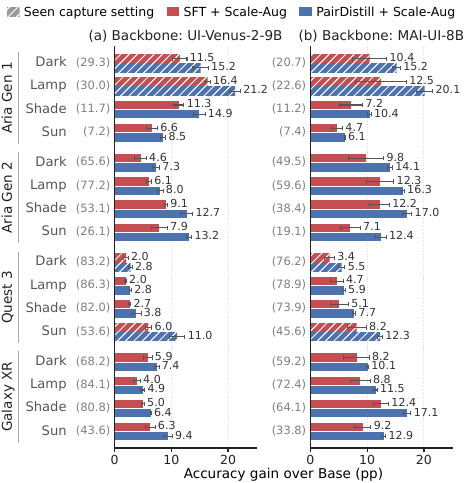}
\caption{\textbf{Gains of SFT and PairDistill over Base across device and lighting combinations.}
Gains are Physical Single-pass accuracy improvements (pp) over Base for (a) UI-Venus-2-9B and (b) MAI-UI-8B, averaged over three training seeds.
Error bars denote the standard deviation across seeds.
Both methods use scale augmentation.
Numbers in parentheses give Base accuracy (\%).
Hatching marks combinations seen in training.}
\label{fig:adaptation_generalization}
\end{figure}

\begin{figure}[t]
\centering
\includegraphics[width=\columnwidth]{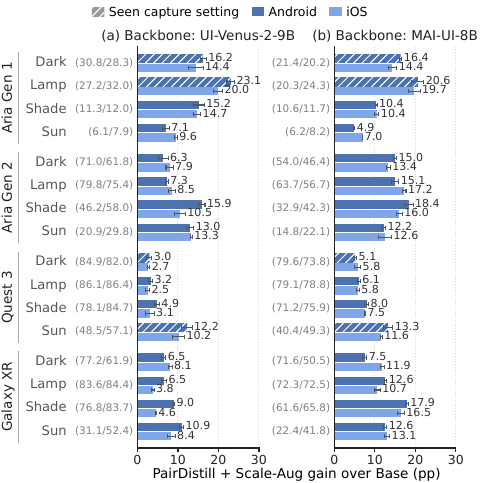}
\caption{\textbf{Gains of PairDistill over Base on Android and iOS.}
Gains are Physical Single-pass accuracy improvements (pp) of PairDistill with scale augmentation over Base for (a) UI-Venus-2-9B and (b) MAI-UI-8B, averaged over three training seeds.
Error bars denote the standard deviation across seeds.
Numbers in parentheses give Base accuracy (\%) on Android and iOS.
Hatching marks the Android subsets of combinations seen in training.
The training data contain no iOS screenshots.}
\label{fig:adaptation_os}
\end{figure}

\subsection{Stage-wise Performance Analysis}
\label{supp:stagewise}

Table~\ref{tab:supp_stagewise} reports the accuracy gaps between consecutive observation stages for the two backbones.
ReGround mainly reduces the gaps before Warped.
It reduces the Clean-to-Resized gap by 5.7 and 8.3 pp on UI-Venus-2-9B and MAI-UI-8B, respectively, and the Resized-to-Warped gap by 1.6 and 6.0 pp, while the Warped-to-Physical gap changes by about 1 pp.
PairDistill instead mainly reduces the Warped-to-Physical gap, by 6.0 and 5.3 pp relative to Base, and also reduces the Resized-to-Warped gap by 1.4 and 5.6 pp.
These two stages together account for 81\% and 84\% of its reduction in the overall Clean-to-Physical gap.
Compared with SFT under the same scale augmentation, the alignment term mainly reduces the Warped-to-Physical gap, by 2.1 and 2.5 pp, while the Clean-to-Resized gap changes by less than 0.4 pp.
Applying ReGround on top of PairDistill further reduces the Clean-to-Resized gap by 4.0 and 6.1 pp, while the Warped-to-Physical gap changes by at most 1.4 pp.
ReGround thus mainly reduces the gaps before Warped, whereas PairDistill mainly reduces the Warped-to-Physical gap, and the two are complementary.

\begin{table}[t]
\centering
\caption{\textbf{Stage-wise accuracy gaps.}
Gaps (pp) between consecutive observation stages: Clean to Resized (C$\to$R), Resized to Warped (R$\to$W), and Warped to Physical (W$\to$P).
ReGround is not applied to Clean, so rows with ReGround use the Single-pass Clean accuracy.
SFT and PairDistill (PD) both use scale augmentation and are averaged over three training seeds.
Gaps are computed from unrounded accuracies.}
\label{tab:supp_stagewise}
\footnotesize
\begin{tabular*}{\linewidth}{@{\extracolsep{\fill}}llccc@{}}
\toprule
Backbone & Setting & C$\to$R & R$\to$W & W$\to$P \\
\midrule
UI-Venus-2-9B & Base            & 8.02  & 3.49 & 29.31 \\
              & + ReGround      & 2.28  & 1.88 & 28.09 \\
              & + SFT           & 6.67  & 2.39 & 25.46 \\
              & + PD            & 6.34  & 2.10 & 23.32 \\
              & + PD + ReGround & 2.33  & 1.05 & 23.24 \\
\midrule
MAI-UI-8B     & Base            & 11.32 & 8.71 & 29.62 \\
              & + ReGround      & 3.02  & 2.68 & 30.72 \\
              & + SFT           & 9.18  & 4.58 & 26.90 \\
              & + PD            & 9.28  & 3.13 & 24.36 \\
              & + PD + ReGround & 3.20  & 1.44 & 25.73 \\
\bottomrule
\end{tabular*}
\end{table}

\subsection{Gains across Capture Conditions}
\label{supp:device_lighting_gains}

Figure~\ref{fig:adaptation_generalization} breaks down the gains of SFT and PairDistill over Base by device and lighting combination.
Averaged over three training seeds, PairDistill achieves larger gains than SFT in all 16 combinations on both backbones, including the 12 combinations absent from training.

\subsection{Transfer from Android to iOS}
\label{supp:os_gains}

Figure~\ref{fig:adaptation_os} breaks down the gains of PairDistill over Base by operating system.
Although the training data contain only Android screenshots, both backbones improve over Base in all 16 combinations on iOS.



\section{Limitations, Ethics, and Dataset Documentation}
\label{supp:limitations_ethics}


\smallskip\noindent\textbf{Limitations.}\quad
GlassGround focuses on single-step grounding in static GUI states and does not cover sequential interaction, action feedback, or complete task execution.
The benchmark is limited to smartphone screens.
By holding GUI content fixed while varying wearable devices and lighting conditions, our controlled capture protocol enables paired comparisons of grounding performance across physical observations.
However, it does not cover dynamic interfaces, other types of physical displays, or the broader range of conditions encountered in everyday use.

\smallskip\noindent\textbf{Ethical considerations.}\quad
This study evaluates GUI grounding models rather than human behavior or performance.
Physical observations were captured by members of the research team following a predefined protocol, without recruiting external participants as research subjects.
We do not analyze individual behavior, physiological characteristics, or other personal attributes.
We use only RGB images from the wearable devices to construct and evaluate the benchmark, without using audio, eye-tracking, or other sensor data.
For these reasons, we did not seek institutional review board (IRB) review.

\smallskip\noindent\textbf{Privacy protection.}\quad
The GUI screenshots in GlassGround were obtained partly from test accounts and partly from personal accounts belonging to members of the research team.
Before public release, we \ifarxiv redacted\else will redact\fi{} non-public personal information, such as names, profile pictures, phone numbers, email addresses, and residential addresses of team members and their contacts.
The corresponding physical images and grounding instructions \ifarxiv were\else will be\fi{} de-identified consistently.
Manual inspection confirmed that no bystanders appeared in indoor captures.
For outdoor captures, we \ifarxiv blurred or redacted\else will blur or redact\fi{} identifiable faces, license plates, and other information that could reveal personal identities.
As a result, the public release \ifarxiv contains\else will contain\fi{} localized differences in some images and task instructions compared with the version used for the evaluations reported in this paper.

\smallskip\noindent\textbf{Dataset documentation and intended use.}\quad
\ifarxiv
We release GlassGround in our repository,
\else
We plan to publicly release GlassGround upon acceptance,
\fi
including clean GUI screenshots, corresponding wearable camera observations, grounding instructions, target region annotations, and human-verified screen corner annotations.
The accompanying documentation will describe the devices, lighting conditions, image correspondences, and evaluation protocols needed to support reproducibility.
GlassGround is intended for research on GUI grounding, visual robustness, and wearable vision understanding.
It is not intended for personal identification or the analysis of individual behavior.

\section{Live Headset Prototype}
\label{supp:prototype_video}

\begin{figure*}[t]
  \centering
  \includegraphics[width=\linewidth]{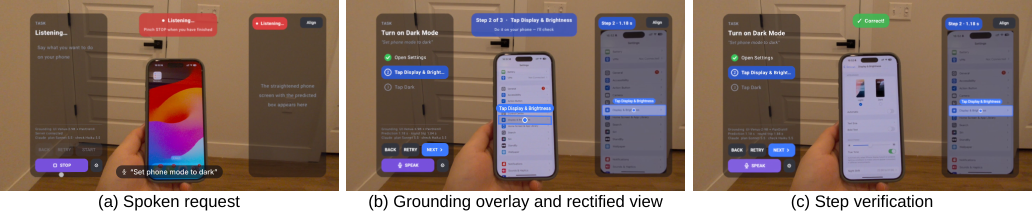}
  \caption{\textbf{Live headset prototype.}
  \ifarxiv
  Frames from the demo video.
  \else
  Frames from the supplementary video.
  \fi
  (a) The user states a task by voice. The subtitle was added in post-production.
  (b) The grounding prediction is rendered as an overlay in the headset's passthrough
  view, aligned with the phone screen. The side panel shows the rectified screen with
  the same prediction. (c) The system detects that the step is complete.}
  \label{fig:supp_prototype}
\end{figure*}

\ifarxiv
The demo video shows a live prototype using a Samsung Galaxy XR headset.
\else
The supplementary video shows a live prototype using a Samsung Galaxy XR headset.
\fi
The user wears the headset and operates a phone step by step, following the overlays
that the assistant renders in real time. The prototype illustrates the application
setting that GlassGround targets. The user states a task by voice. The assistant has
no access to the phone's screenshots or interface state. It observes the screen only
through the headset camera and highlights the element to operate at each step. In the
video, the user completes three tasks in the iOS Settings app: enabling Dark Mode,
setting brightness to maximum, and setting the largest text size
(Figure~\ref{fig:supp_prototype}). Guidance in this setting depends on localizing the
target element in the camera view, which is the capability GlassGround evaluates.

\smallskip\noindent\textbf{Pipeline.}\quad
The headset handles the interface, camera capture, and phone tracking. Speech
transcription and grounding run on a remote GPU server, and Claude models are called
through their API. The spoken request is transcribed with Whisper large-v3-turbo.
Claude Sonnet 5.5 then decomposes it into steps and writes a grounding instruction and
a completion condition for each step. For each step, the grounding model
(UI-Venus-2-9B + PairDistill) takes the grounding instruction and predicts the target
once on the original headset camera frame, as in Single-pass inference. The screen
corner detector from Sec.~\ref{sec:gt_transfer} separately locates the phone screen in
the same frame, and the resulting homography converts the prediction to screen
coordinates. A headset-side version of the same detector tracks the phone in real
time. The overlay is continuously re-projected onto the live passthrough view
according to the tracking result, so it follows the phone as it moves. A side panel
shows the rectified screen from the captured frame together with the prediction.
While a step is active, the system repeatedly captures a frame and rectifies the
screen, and Claude Haiku 5.5 checks it against the completion condition. Once the
condition is met, the system advances to the next step automatically.

\smallskip\noindent\textbf{Latency.}\quad
In this setting, the user waits for the overlay before acting at each step, so
per-step latency directly affects how smoothly the guidance proceeds. This is also why
Sec.~\ref{sec:evaluation_setup} considers inference efficiency when selecting
closed-source baselines. The prototype calls the grounding model once per step. Across
the six steps in the video, grounding takes 1.17 to 1.18\,s on the server. The round
trip from sending the grounding request to receiving the result takes 1.56 to
1.84\,s. It includes upload, server processing, and download, and excludes the
preceding frame capture and the subsequent rendering.

\smallskip\noindent\textbf{Live operation and recording.}\quad
Grounding, overlay rendering, and step verification in the video all run live while
the user operates the phone. They are not produced by offline processing of a recorded
video. The video frames are the headset's screen recording from a single continuous
take, with no cuts between the three tasks. The overlays and panels are what the
prototype rendered live in the headset and what the user saw, although the recorded
field of view is slightly narrower than the view when wearing the headset. The user
completed each step by following these cues. Apart from trimming the clip,
post-production only affected the audio and subtitles. In this recording, the
headset's screen capture contains no sound while the application is using the
microphone. The speech in the video therefore comes from the same recording that the
application saved for transcription,
voice-altered for \ifarxiv privacy\else anonymity\fi{} and placed
back at its original timing. The subtitles show the application's transcription and were overlaid afterwards. Ambient sound was removed. The video illustrates the application setting and is not an evaluation result.

\onecolumn

\section*{Grounding Prompt Templates}
The following prompt templates were used for grounding inference.
The placeholders \texttt{\{instr\}} and \texttt{\{instruction\}} denote the current task instruction.
For P1 and P2, the template is a system prompt and the user message contains the task instruction.
The other templates are user prompts or structured requests.

\begin{promptbox}{P1: MAI-UI-8B (System Prompt)}
You are a GUI grounding agent.

## Task

Given a screenshot and the user's grounding instruction. Your task is to accurately locate a UI element based on the user's instructions.

First, you should carefully examine the screenshot and analyze the user's instructions,  translate the user's instruction into a effective reasoning process, and then provide the final coordinate.

## Output Format

Return a json object with a reasoning process in <grounding_think></grounding_think> tags, a [x,y] format coordinate within <answer></answer> XML tags:

<grounding_think>...</grounding_think>

<answer>
{"coordinate": [x,y]}
</answer>

## Input instruction
\end{promptbox}

\begin{promptbox}{P2: GUI-Owl, KV-Ground, and Indeed-UI (System Prompt)}
# Tools

You may call one or more functions to assist with the user query.

You are provided with function signatures within <tools></tools> XML tags:

<tools>
{"type": "function", "function": {"name": "computer_use", "description": "Use a mouse to interact with a computer.

* The screen's resolution is 1000x1000.

* Make sure to click any buttons, links, icons, etc with the cursor tip in the center of the element. Don't click boxes on their edges unless asked.

* don't use any other computer use tool like type, key, scroll, left_click_drag and so on.

* you can only use the left_click and mouse_move action to interact with the computer. if you can't find the element, you should terminate the task and report the failure.", "parameters": {"properties": {"action": {"description": "The action to perform. The available actions are:

* `mouse_move`: Move the cursor to a specified (x, y) pixel coordinate on the screen.

* `left_click`: Click the left mouse button with coordinate (x, y) pixel coordinate on the screen.", "enum": ["mouse_move", "left_click"], "type": "string"}, "coordinate": {"description": "(x, y): The x (pixels from the left edge) and y (pixels from the top edge) coordinates to move the mouse to. Required only by `action=mouse_move` and `action=left_click`.", "type": "array"}}, "required": ["action"], "type": "object"}}}
</tools>

For each function call, return a json object with function name and arguments within <tool_call></tool_call> XML tags:

<tool_call>
{"name": <function-name>, "arguments": <args-json-object>}
</tool_call>
\end{promptbox}

\begin{promptbox}{P3: UI-Venus-2-9B and UI-Mate-27B (User Prompt)}
Outline the position corresponding to the instruction: {instr}. The output should be only [x1,y1,x2,y2].
\end{promptbox}

\begin{promptbox}{P4: Holo3.1, Claude Sonnet 5.5, and GPT-6 Luna (User Prompt)}
Localize an element on the GUI image according to the provided target and output a click position.

* You must output a valid JSON following the format: {"x": int, "y": int}, where x and y are normalized between 0 and 1000.

{instr}
\end{promptbox}

\begin{promptbox}{P5: Muse Glimmer-30B (User Prompt)}
Given the UI screenshot, output the click coordinates for: {instr}. Respond with only [x, y] coordinates normalized 0-1000.
\end{promptbox}

\begin{promptbox}{P6: Qwen3.8-27B (User Prompt)}
Look at this UI screenshot. Point at the UI element matching: {instr}. Output only the pixel coordinates as [x, y], normalized 0-1000.
\end{promptbox}

\begin{promptbox}{P7: Jev 1.14 (Structured Request)}
State:
Screenshot of a user interface:

Question:
{
  "target": {
    "type": "box",
    "instructions": "In the screenshot, the UI element to use for this task: {instruction}"
  }
}
\end{promptbox}


\end{document}